\documentclass[lettersize,journal]{IEEEtran}
\usepackage{amsmath,amsfonts}
\usepackage{algorithmic}
\usepackage{algorithm}
\usepackage{array}
\usepackage{multirow}
\usepackage[caption=false,font=normalsize,labelfont=sf,textfont=sf]{subfig}
\usepackage{textcomp}
\usepackage{stfloats}
\usepackage{url}
\usepackage{verbatim}
\usepackage{graphicx}
\usepackage{cite}
\begin{document}

\title{LookupForensics: A Large-Scale Multi-Task Dataset for Multi-Phase Image-Based Fact Verification}

\author{Shuhan Cui,~\IEEEmembership{Member,~IEEE,} 
Huy H. Nguyen,~\IEEEmembership{Member,~IEEE,} 
Trung-Nghia Le,~\IEEEmembership{Member,~IEEE,} \\
Chun-Shien Lu,~\IEEEmembership{Member,~IEEE,} 
Isao Echizen,~\IEEEmembership{Senior Member,~IEEE,}

\thanks{This work was partially supported by JSPS KAKENHI Grants JP21H04907 and JP24H00732, and by JST CREST Grants JPMJCR18A6 and JPMJCR20D3 including AIP challenge program, and by JST AIP Acceleration Grant JPMJCR24U3 Japan.}%
\thanks{Shuhan Cui is with the Graduate School of Information Science and Technology, University of Tokyo, Tokyo 113-0033, Japan (e-mail: syokan@g.ecc.u-tokyo.ac.jp).}
\thanks{Huy H. Nguyen is with the Echizen Laboratory, National Institute of Informatics, Tokyo 101-0003, Japan (e-mail: nhhuy@nii.ac.jp).}
\thanks{Trung-Nghia Le is with the University of Science, VNU-HCM, Ho Chi Minh City, Vietnam, and also with the Vietnam National University, Ho Chi Minh City, Vietnam (orcid: 0000-0002-7363-2610, e-mail: ltnghia@fit.hcmus.edu.vn).}
\thanks{Chun-Shien Lu is with the Institute of Information Science, Academia Sinica, Taipei 115201, Taiwan (e-mail: lcs@iis.sinica.edu.tw).}
\thanks{Isao Echizen is with the Graduate School of Information Science and Technology, University of Tokyo, Tokyo 113-0033, Japan, and also with the National Institute of Informatics, Tokyo 101-0003, Japan (e-mail: iechizen@nii.ac.jp).}
\thanks{Digital Object Identifier .}
}

\markboth{IEEE TRANSACTIONS ON INFORMATION FORENSICS AND SECURITY,~Vol.~, No.~, ~}%
{CUI \MakeLowercase{\textit{et al.}}: LookupForensics: A Large-Scale Multi-Task Dataset for Multi-Phase Image-Based Fact Verification}

\maketitle

\begin{abstract}
Amid the proliferation of forged images, notably the tsunami of deepfake content, extensive research has been conducted on using artificial intelligence (AI) to identify forged content in the face of continuing advancements in counterfeiting technologies. We have investigated the use of AI to provide the original authentic image after deepfake detection, which we believe is a reliable and persuasive solution. We call this \textit{``image-based automated fact verification,''} a name that originated from a text-based fact-checking system used by journalists. We have developed a two-phase open framework that integrates detection and retrieval components. Additionally, inspired by a dataset proposed by Meta Fundamental AI Research, we further constructed a large-scale dataset that is specifically designed for this task. This dataset simulates real-world conditions and includes both content-preserving and content-aware manipulations that present a range of difficulty levels and have potential for ongoing research. This multi-task dataset is fully annotated, enabling it to be utilized for sub-tasks within the forgery identification and fact retrieval domains. This paper makes two main contributions: (1) We introduce a new task, `\textit{`image-based automated fact verification,''} and present a novel two-phase open framework combining \textit{``forgery identification''} and \textit{``fact retrieval.''} (2) We present a large-scale dataset tailored for this new task that features various hand-crafted image edits and machine learning-driven manipulations, with extensive annotations suitable for various sub-tasks. Extensive experimental results validate its practicality for fact verification research and clarify its difficulty levels for various sub-tasks.
\end{abstract}


\section{Introduction}
\label{sec:introduction}
Forgery techniques continue to rapidly evolve, especially with the emergence of deepfake generators like generative adversarial networks (GANs) and diffusion models, which enable the creation of highly realistic forgeries that can deceive image forgery detection systems and even humans. Moreover, conventional forgery techniques, such as image splicing, copy-move, object removal, and colorization \cite{Faceapp, inc_2020, kim2018deep, learning_2020, zakharov2019few} have also advanced with the help of deep learning. These developments have made it increasingly easy for anyone to create convincing forgery images.

\begin{figure*}[t]
    \begin{center} 
        \includegraphics[width=1.0\linewidth]{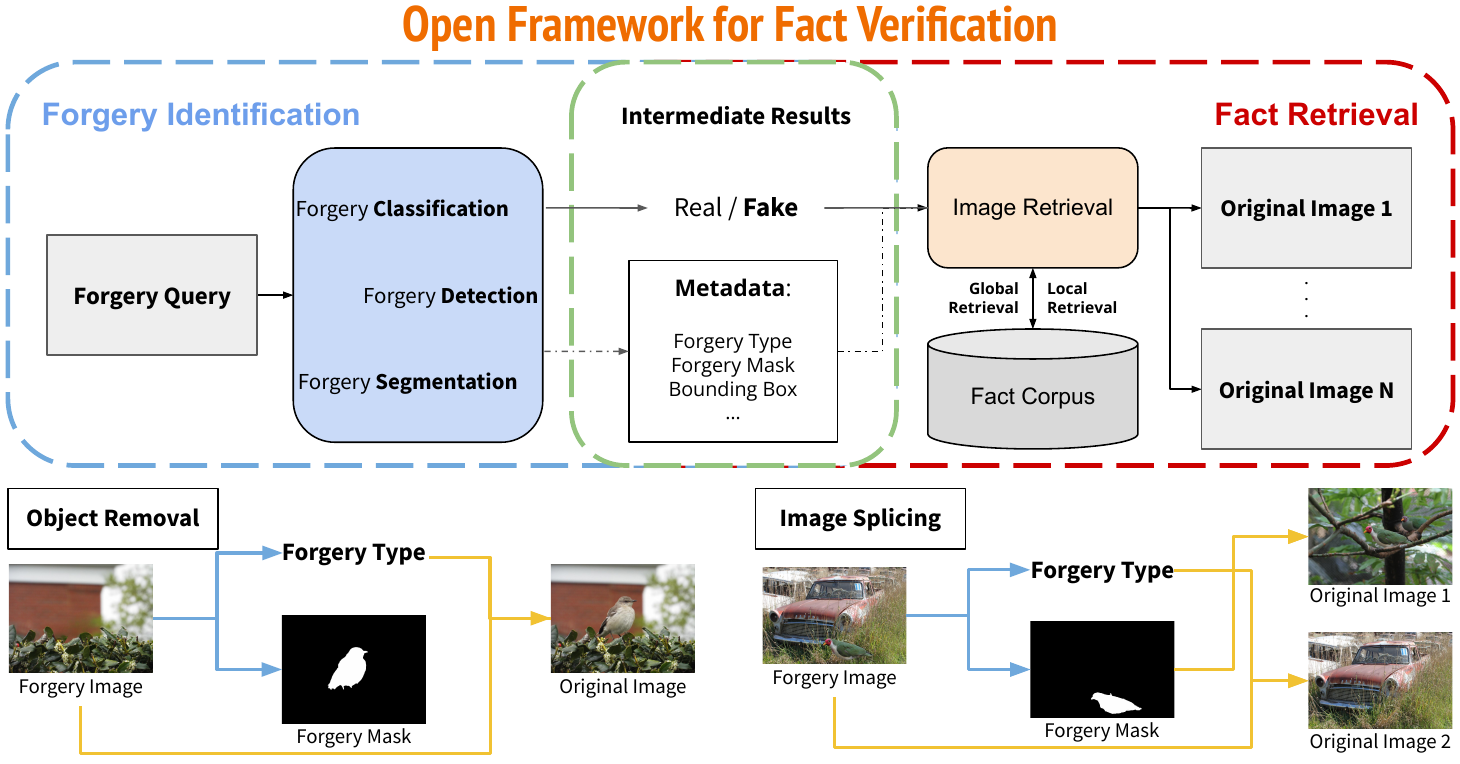}
    \end{center}
    \caption{Upper half illustrates pipeline of our proposed open framework for image-based fact verification: It comprises two phases: forgery identification and fact retrieval. The modules therein can be replaced with almost any open-source toolbox. Lower half shows examples of two specific forgery types.}
    \label{fig:framework}
\end{figure*}

Despite their remarkable capabilities, forgery techniques have gained notoriety for being used unethically and maliciously, such as for spreading misleading information, creating fake pornographic content for extortion, and fabricating inflammatory political statements. Moreover, the widespread adoption of intelligent devices with high-quality cameras and extensive storage has accelerated the dissemination of all types of information, thereby amplifying the effect of forged content. To curb the proliferation of fake information, numerous forgery detection methods based on artificial intelligence (AI)\cite{li2020face, masi2020two, matern2019exploiting, qian2020thinking, wu2020sstnet} have been devised. Some of them treat forgery detection as a binary classification problem, utilizing backbone networks to extract global features from suspect images and using binary classifiers to distinguish between real and fake content. Other methods \cite{le2021openforensics, hsu2007image, mp2017copy, katircioglu2021self} are aimed at precisely detecting the forged positions of an image. Despite the effectiveness of these AI-based detection methods, incomplete and faulty data sources and frequently exposed errors have significantly eroded public trust in AI. Therefore, instead of simply providing a detection result, a more reliable and persuasive approach is to provide the original authentic images alongside the detection result, which we call \textbf{image-based automated fact verification}.

The concept of \textbf{fact verification} originates from the text-based fact-checking systems utilized by journalists. These validation systems meticulously assess the credibility of factual claims, statements, or information through thorough investigation and analysis, which closely aligns with our approach of authenticating images and identifying their original counterparts. Prevailing fact verification systems \cite{graves2018understanding, guo2022survey, walter2020fact} rely heavily on manual interventions and are primarily aimed at texts rather than images. Therefore, image-based automated fact verification represents a novel endeavor in this domain. For this new task, we introduce a \textbf{two-phase open framework} that integrates the forgery identification phase with a retrieval phase to detect forgeries and retrieve originals, as illustrated in Figure \ref{fig:framework}. This framework ensures that detection results contribute to effective retrieval while retrieval results assist in optimizing detection results and enhancing interpretability. 

Additionally, to bolster research in image-based fact verification, we further constructed a \textbf{large-scale}  specifically designed dataset inspired by the Image Similarity Challenge 2021 (ISC2021) \cite{douze20212021} benchmark proposed by Meta Fundamental AI Research. The ISC2021 benchmark is aimed at enhancing the traceability of images by focusing on a task called ``image copy detection (ICD)'' for determining if an input image is a manipulated version of an image from a database, which closely aligns with our objectives in the retrieval phase. We thus constructed a dataset mimicking that used in ISC2021. The dataset simulates real-world conditions and supports various image manipulations, including automated, manual, and machine-learning-driven manipulations. Our dataset also incorporates a significant proportion of unrelated distractors in its reference and query sets. Whereas the ISC2021 dataset was aimed more at image traceability than authenticity, ours is aimed at both traceability and authenticity, which leads to some differences. The ISC2021 dataset primarily features traditional content-preserving manipulations, whereas our dataset features deep-learning-generated content-aware manipulations that can substantially alter image content while maintaining realism.

Moreover, our multi-task dataset is thoroughly annotated, enabling it to be used across different forgery types for various sub-tasks, such as forgery detection and classification. Extensive evaluations demonstrated that it is suitable for both image-based automated fact verification and various sub-tasks in forgery identification and fact retrieval.

\begin{figure*}[t]
    \begin{center} 
        \includegraphics[width=0.96\linewidth]{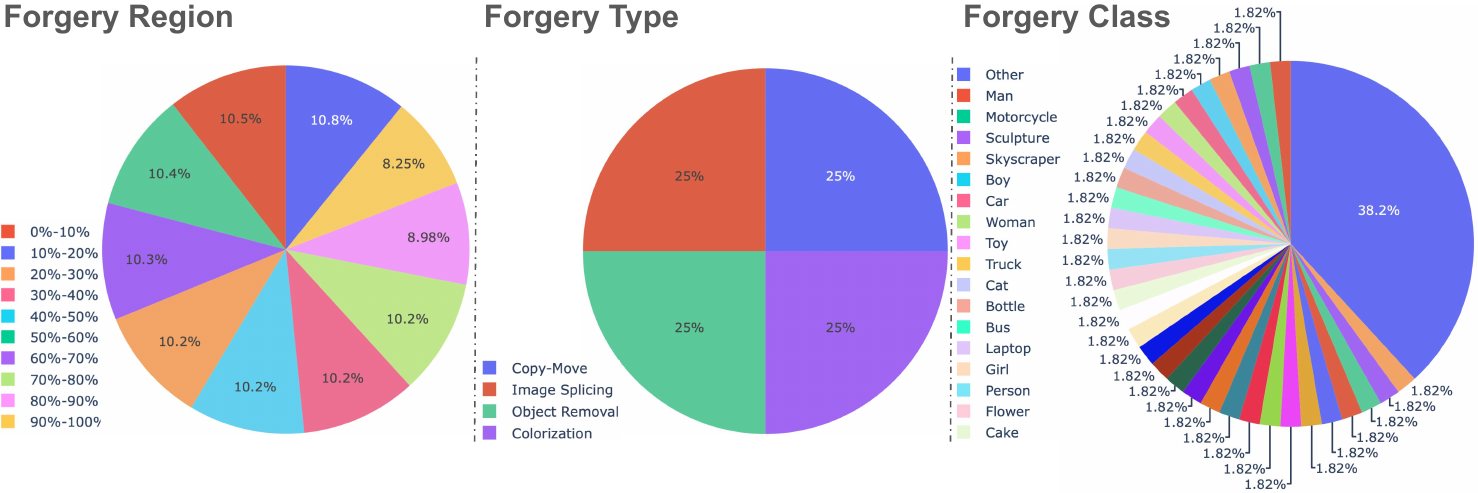}
    \end{center}
    \caption{Distribution of the proportion of forgery region, forgery type, and forgery object class in our dataset.}
    \label{fig:datasetstatistics}
\end{figure*}

This paper makes two contributions. (1) We introduce a two-phase open framework that addresses the challenge of image-based automated fact verification that integrates forgery identification and fact retrieval. (2) We present a large-scale dataset tailored for this new task. Built on Google's Open Images Dataset \cite{kuznetsova2020open}, our dataset encompasses various forgery types with varying difficulty levels, including automated transformations, hand-crafted image edits, and machine learning-driven manipulations. It also features diverse annotations serving various research purposes. The distributions of this dataset in terms of forgery region, forgery type, and forgery object class are shown in Figure \ref{fig:datasetstatistics}. Comprehensive evaluations demonstrated that our dataset is well studied for advancing image-based fact verification and related sub-tasks. Open access to the entire dataset, evaluation toolkit, and trained models will be provided upon publication.

The remainder of this paper is organized as follows: In Section \ref{sec:relatedwork}, we explain background information, including general fact verification frameworks, image forgery detection, ICD, and related datasets. Section \ref{sec:dataset} describes our framework and dataset for image-based fact verification. Section \ref{sec:evaluation} presents the results of experimental evaluation on different components of the framework. Section \ref{sec:discussion} discusses various questions arising from this research, illustrated with experimental comparisons. Finally, concluding remarks and future plans are given in Section \ref{sec:conclusion}.

\section{Related Work}
\label{sec:relatedwork}
In this section, we first discuss the general framework for fact verification in Section \ref{sec:FactVerification}, followed by explanations of the image forgery detection and ICD tasks in Sections \ref{sec:ImageForgeryDetection} and \ref{sec:ImageCopyDetection}, respectively. We conclude with an overview of relevant datasets in Section \ref{sec:relateddatasets}.

\subsection{Fact Verification}
\label{sec:FactVerification}

Fact verification, also known as fact-checking, is a systematic process aimed at determining the \textbf{accuracy} and \textbf{credibility} of claims, statements, or information circulating in various media. It plays a crucial role in combating misinformation and disinformation by providing \textbf{evidence-based assessments} of the validity of claims. As outlined by Graves \cite{graves2018understanding, guo2022survey, walter2020fact}, fact-checking typically involves three overlapping phases: identification, verification, and correction. In the \textbf{identification} phase, fact-checkers identify claims or statements to be verified. These claims can originate from various sources, such as news articles, social media posts, or public speeches. During the \textbf{verification} phase, fact-checkers conduct thorough research by consulting authoritative sources, examining official documents, and sometimes interviewing experts. This meticulous analysis and cross-referencing of information helps the fact-checkers to assess the accuracy and credibility of the claim. Finally, in the \textbf{correction} phase, the fact-checkers present their findings to the public through detailed reports or articles, ensuring that accurate information is disseminated.

While image-based fact verification remains relatively unexplored, we can leverage insights from traditional text-based frameworks. Similarly, in the identification phase, state-of-the-art (SOTA) forgery detection networks can replace human experts to provide robust detection. During the verification phase, cross-referencing suspicious images against a repository of reference images (with confirmed authenticity) enables retrieval of the corresponding original sources. In the correction phase, presenting the original images alongside the detection results enhances the credibility of the fact verification process and mitigates concerns about diminishing trust in the ability of AI to combat misinformation and disinformation. In this paper, we focus on the \textbf{first two phases} and explore the feasibility of image-based fact verification by presenting a novel two-phase open framework.
\subsection{Image Forgery Detection}\label{sec:ImageForgeryDetection}

According to the detection approach, there are two general types of forgery techniques: forgery-independent and forgery-dependent. Forgery-dependent techniques target specific features that are more susceptible to forgery and include copy-move, image splicing, and object removal. In contrast, forgery-independent techniques typically target intrinsic and difficult-to-replicate characteristics, such as color.

\textbf{Copy-Move}, one of the most straightforward techniques, involves duplication of one or more portions of a digital image and inserting them elsewhere within the same image, aiming to deceive viewers by creating the illusion of multiple instances of the same object within the image. \textbf{Image splicing} involves manipulation of digital images by combining elements from various sources, enabling perpetrators to add or remove objects, alter backgrounds, or fabricate events within the image. \textbf{Object removal} involves deliberate alteration of digital images to remove specific objects or entities from the image while maintaining the overall context of the image, which is often done to conceal or erase unwanted elements from photographs, such as people, logos, or text. \textbf{Colorization} involves the digital modification of the color distribution of an image to change its overall appearance or convey a different mood or message, which can enhance or distort the visual content and thereby lead to potential misinterpretations or misrepresentations. 

Existing detection methods generally take one of two approaches: active or passive approach. Ones taking the \textbf{active approach} are designed for real-time detection and make use of previously embedded hidden information in digital media for use in asserting ownership rights, tracking content distribution, and deterring unauthorized use or manipulation. Prominent methods include digital watermarking \cite{arnold2003techniques, koptyra2020imagechain} and digital steganography \cite{nikolaidis1996copyright, gjosteen2018practical}. Ones taking the \textbf{passive approach} do not use image \textbf{preprocessing} but instead rely on the premise that tampering with the original image introduces inconsistencies in its statistical properties. Passive methods thus use statistical analysis, signal processing, or machine learning to detect subtle discrepancies that betray the authenticity of visual content. Among the passive methods, forgery detection methods for copy-move have been the most studied.

Shivakumar and Baboo suggested using speeded-up robust features (SURF) for detecting copy-move image forgery by using the KD-Tree algorithm to extract and match SURF key points \cite{shivakumar2011detection}. With the development of \textbf{machine learning}, Chen et al. were the first to use convolutional neural networks (CNNs) in median filtering image forensics; they replaced the first layer with a filter layer designed to output the input image's median filtering residual (MFR) \cite{chen2015median}. Furthermore, the use of convolutional and pooling layers enabled multiple features to be used for further classification. Shi et al. developed a unified framework, a dual-domain-based CNN (D-CNN), for detecting copy-move forgeries \cite{shi2018image}. It comprises two sub-networks, a sub-SCNN and a sub-FCNN, that work together to identify the tampered regions in a transfer learning way.

\begin{table*}[t]
\centering
\caption{Basic information about forgery datasets. ``Spl.," ``Cop.," ``Rem.," and ``Col." represent image splicing, copy-move, object removal, and colorization, respectively. ``GT Types'' represents the type(s) of ground truth provided in the dataset, including forgery mask (Mask), forgery class (Class), forgery instances (Instance), and forgery bounding box (Bounding Box).}
\label{table:ForgeryDataset}
\begin{tabular}{lrrrrrr}
\hline
\textbf{Dataset}    & \textbf{Year} & \textbf{GT Types} & \textbf{Forgery types}                     & \textbf{No. Real} & \textbf{No. Forged} & \textbf{Dimensions}     \\ \hline
Columbia Gray \cite{ng2004data} & 2004          & -                        & Spl.                                   & 933 & 912                     & 128 x 128               \\
Columbia Color \cite{hsu2006detecting} & 2006          & Mask                        & Spl.                                   & 183 & 180                     & 757 x 568 to 1,152 x 768 \\
MICC F2000 \cite{amerini2011sift} & 2011          & -                        & Cop.                                  & 1,300 & 700                    & 2,048 x 1,536             \\
FAU/Manip \cite{christlein2012evaluation} & 2012          & Mask                        & Cop.                                  & 48 & 87                       & Average 3,000 × 2,300       \\
VIPP Synth \cite{bianchi2012image} & 2012          & Mask                        & Spl.                                   & 4,800 & 4,800                   & Various                 \\
CASIA v1.0 \cite{dong2013casia} & 2013          & -                        & Spl., Cop.                        & 800 & 921                     & 384 x 256               \\
CASIA v2.0 \cite{dong2013casia} & 2013          & -                        & Spl., Cop.                        & 7,491 & 5,123                   & 240 x 160 to 900 x 600  \\
Carvalho \cite{de2013exposing} & 2013          & Mask                        & Spl.                                   & 100 & 100                     & 2,048 x 1,536           \\
CoMoFoD \cite{tralic2013comofod} & 2013          & Mask                        & Cop.                                  & 260 & 260                     & Various                 \\
GRIP \cite{cozzolino2015efficient} & 2015          & Mask                        & Cop.                                  & 80 & 80                       & 768×1,024                \\
Wild Web \cite{zampoglou2015detecting} & 2015          & Mask                        & Spl.                                   & 90 & 9,657                     & Various                 \\
Realistic Tampering \cite{korus2016multi} & 2016          & Mask                        & Rem.                                   & 220 & 220                     & 1,920 x 1,080           \\
COVERAGE \cite{wen2016coverage} & 2016          & Mask                        & Cop.                                  & 100 & 100                     & Average 400 x 486       \\
In the Wild \cite{huh2018fighting} & 2018          & Mask                        & Spl.                                   & - & 201                       & Various                 \\
PS-Battles \cite{heller2018ps} & 2018          & -                        & Spl., Cop., Rem.               & 11,142 & 102,028                & Various                 \\
Fantastic Reality \cite{kniaz2019point} & 2019          & Mask, Instance, Class    & Spl.                                   & 16,000 & 16,000               & Various                 \\
DEFACTO \cite{mahfoudi2019defacto} & 2019          & Mask                        & Spl., Cop., Rem.               & - & 229,000                    & Various                 \\
IMD2020 \cite{novozamsky2020imd2020} & 2020          & Mask                        & Spl., Cop., Rem.               & 35,000 & 35,000                 & Various                 \\ \hline
\textbf{Ours}                & \textbf{2024}          & \textbf{Mask, Class, Bounding Box}   & \textbf{Spl., Cop., Rem., Col.} & \textbf{40,000} & \textbf{40,000}               & \textbf{Various}                 \\ \hline
\end{tabular}
\end{table*}

\subsection{Image Copy Detection}
\label{sec:ImageCopyDetection}

ICD, also known as image plagiarism detection, involves determining whether an image has been replicated. Since its inception, ICD has been a critical research area for identifying image forgery. Early methods relied on visual features such as color histograms, texture descriptors, or SIFT (scale-invariant feature transform) features to compare a query image with a set of reference images to identify potential matches. However, as noted by Douze et al. \cite{douze20212021}, the task of ICD was generally regarded as resolved after years of research. SOTA approaches have proficiently handled existing datasets, leading to a decline in the creation of new datasets and methods. Recent research \cite{douze20212021}, however, has rekindled interest in this field by introducing new challenges within the task.

ISC2021 served as a platform to foster the creation of effective methods for ICD. Participants were provided with a large-scale dataset comprising various images and were tasked with devising algorithms capable of accurately and efficiently retrieving original images from the reference set corresponding to a given query image. The DISC21 benchmark introduced two novel aspects that address limitations in previous research. The first aspect is \textbf{scale}. DISC21's authors argued that ICD systems must be designed to operate at a massive scale, given that platforms like Facebook contend with billions of uploaded images daily. In real-world scenarios, most uploaded images have no corresponding original images with which to match, so the system's efficiency hinges on its ability to handle these non-match queries. Moreover, since the images to be detected are low-prevalence targets (i.e., a ``needle in a haystack''), the weight of false positives increases rapidly as the dataset is enlarged. Hence, the practicality of focusing solely on high-confidence matches is essential. Consequently, the DISC21 benchmark highlighted the need for a large-scale dataset featuring significant percentages of unmatched distractors in the query and reference sets. The second aspect is \textbf{new attacks}. ICD systems are confronting increasingly sophisticated manipulations due to continuing advancements in image manipulation techniques. Users have numerous options for editing images to be posted on social media, and the prevalence of mobile devices has led to widespread media sharing through mobile screenshots and captures. Therefore, the DISC21 benchmark incorporates extensive transformations in the queries, simulating real-world conditions. Additionally, a competition organized beyond the dataset creation drew over 200 participants, significantly advancing research in ICD.

ISC2021 inspired our construction of a new large-scale dataset for image-based automated fact verification. This dataset simulates real-world conditions and includes various image manipulations, ranging from automated and manual edits to machine learning-driven transformations. By supporting traceability and authenticity, our dataset presents unique challenges and opportunities for advancing research in forgery detection and fact verification.

\subsection{Related Datasets}
\label{sec:relateddatasets}

Since our new dataset is tailored to image-based fact verification by simultaneously addressing forgery detection and ICD, we briefly discuss relevant datasets from these two aspects: forgery-related datasets and ICD-related datasets.

\subsubsection{Datasets in Image Forgery Detection}

Table \ref{table:ForgeryDataset} summarizes the basic information of related image forgery datasets. \textbf{Copy-Move} is one of the most common forms of forgery. The MICC-F2200 dataset \cite{amerini2011sift}, the oldest and most widely used dataset in this domain \cite{chihaoui2014copy, uliyan2016image}, consists of $1300$ real images and $700$ fake ones. FAU's image manipulation dataset \cite{christlein2012evaluation} includes $48$ base images with extracted segments from these images. The CASIA dataset \cite{dong2013casia} focuses on copy-move and image splicing, with CASIA v1.0 containing $800$ authentic and $921$ spliced images, and CASIA v2.0 containing $7,491$ authentic and $5,123$ tampered images. The CoMoFoD dataset \cite{tralic2013comofod}, tailored for copy-move forgeries, comprises $260$ manipulated images of two different sizes. The images in the COVERAGE dataset \cite{wen2016coverage} were carefully selected to increase robustness by incorporating various types of tampering.

\textbf{Image splicing} is another widely researched forgery type. The Columbia Gray dataset \cite{ng2004data}, the oldest in this category, consists of $933$ original images and $912$ forgery images. Another variant, Columbia Color \cite{hsu2006detecting}, contains $183$ authentic and $180$ forged images. The Carvalho dataset \cite{de2013exposing} includes a subset from the IEEE Image Forensics Challenge. The WildWeb dataset \cite{zampoglou2015detecting} comprises $9657$ fake images from the Internet that were compressed before uploading. The In the Wild dataset \cite{huh2018fighting} consists of $201$ forgery images that are realistic images from the wild. The Fantastic Reality dataset \cite{kniaz2019point} includes large-scale manipulated images, with $16,000$ real and $16,000$ fake images featuring pixel-level annotations of manipulated areas. 

\textbf{Object removal} manipulation is less frequently studied. The Realistic Tampering dataset \cite{korus2016multi} is a classic example, containing $220$ realistic forgeries created using modern photo editing software.
 
For datasets with \textbf{multiple forgery types}, significant contributions include the DEFACTO dataset \cite{mahfoudi2019defacto}, which has over $200,000$ images with authentic manipulations encompassing image splicing, copy-move, removal, and face morphing. The PS-Battles dataset \cite{heller2018ps} contains $102,028$ images grouped into $11,142$ subsets, each featuring the original image alongside numerous manipulated versions. The IMD2020 dataset \cite{novozamsky2020imd2020} includes $35,000$ images sourced from $2,322$ distinct camera models. 

Current forgery datasets typically focus on a single forgery type with relatively simplistic annotations, limiting the development of versatile forgery detection techniques. Additionally, many of these datasets are relatively small. In contrast, our proposed forgery dataset encompasses a wide range of forgery types with extensive annotations, drawing from various domains to enhance generalization capabilities, as shown in Table \ref{table:ForgeryDataset}.

\subsubsection{Datasets for Image Copy Detection}

\begin{table}
\caption{Datasets related to ours. ``CD," ``ILR," and ``FR" represent image copy detection, instance-level recognition, and fact retrieval. Query means retrievable queries. QueryD means distractor queries.}
\label{table:CopyDetectionDataset}
\centering
\begin{tabular}{lccccc}
\hline
\multirow{2}{*}{Dataset} & \multirow{2}{*}{Year} & \multirow{2}{*}{Task} & \multicolumn{3}{c}{Dataset size}                                   \\ \cline{4-6} 
                         &                       &                       & Queries              & QueryD               & Reference            \\ 
\hline
Copydays \cite{douze2009evaluation} & 2009 & CD   & 157     & -           & 3k        \\
Instre \cite{wang2015instre}   & 2015 & ILR  & 1250    & -           & 27k       \\
GLDv1 \cite{noh2017large}    & 2017 & ILR  & 1.3k    & 117k        & 1.2M/1.1M \\
ROxford \cite{radenovic2018revisiting}  & 2018 & ILR  & 70      & -           & 5k (+1M)  \\
RParis \cite{radenovic2018revisiting}   & 2018 & ILR  & 70      & -           & 6k (+1M)  \\
GLDv2 \cite{weyand2020google}    & 2020 & ILR  & 1.3k    & 117k        & 4.1M/762k \\
DISC21 \cite{douze20212021}   & 2021 & CD   & 20k     & 80k         & 1M        \\ \hline
\textbf{Ours}     & \textbf{2024} & \textbf{FR}   &     \textbf{20k}    &       \textbf{60k}     &   \textbf{800k}       \\ 
\hline
\end{tabular}
\end{table} 

Table \ref{table:CopyDetectionDataset} summarizes the basic information of datasets related to fact retrieval. Besides copy detection (CD), we also compare the datasets related to instance-level recognition (ILR), a concept closely related to the definition of image similarity. Previous approaches to ICD often relied on \textbf{proprietary} datasets for evaluation, although there are publicly accessible datasets such as the relatively small Copydays dataset \cite{douze2009evaluation}. Certain datasets are tailored for specific tasks, such as logo recognition (e.g., the BelgaLogos dataset \cite{joly2009logo}). Before the ISC2021 dataset was created, large-scale datasets for copy detection were rare and faced significant challenges, particularly the lack of user-generated edited copies. Datasets for ILR exist on a much grander scale, often using crowd-sourced labels \cite{weyand2020google}, though they may include some noise. Our dataset construction follows the ISC2021 model, incorporating similar methods to simulate real-world conditions and support diverse manipulations. In short, our dataset surpasses existing datasets in terms of scale, complexity, and dimensionality, making it genuinely \textbf{large-scale}.

\section{Framework and Dataset for Image-based\\ Fact Verification}
\label{sec:dataset}
To achieve image-based fact verification, we need to address several key issues. Our focus is on designing the identification and verification components of a typical fact verification process. In this context, we have developed a multi-phase framework and constructed a multi-task dataset.

\subsection{Framework}

Our open two-phase framework is designed to address the challenge of image-based fact verification. As illustrated in Figure \ref{fig:framework}, it consists of two primary phases: \textbf{forgery identification} and \textbf{fact retrieval}. Unlike previous forgery detection methods, our framework enables the simultaneous retrieval of original images. Additionally, in contrast to traditional fact verification methods \cite{graves2018understanding, guo2022survey, walter2020fact}, our approach is fully automated and represents the first research effort in image-based fact verification.

\subsubsection{Forgery Identification}

The first phase of our framework focuses on forgery identification, wherein each input query undergoes a comprehensive analysis through forgery localization and classification. This phase involves: determining whether an image is authentic or forged (binary classification) and, if the image is identified as forged, identifying the specific type of forgery (forgery classification), and using detection models trained on our dataset to identify forgery bounding boxes or areas on the basis of the identified forgery type (location detection).

This systematic approach ensures thorough investigation and analysis, enabling the system to keep pace with improvements in forgery techniques.

\subsubsection{Fact Retrieval}

\begin{figure}[t!]
    \begin{center} 
        \includegraphics[width=1.0\linewidth]{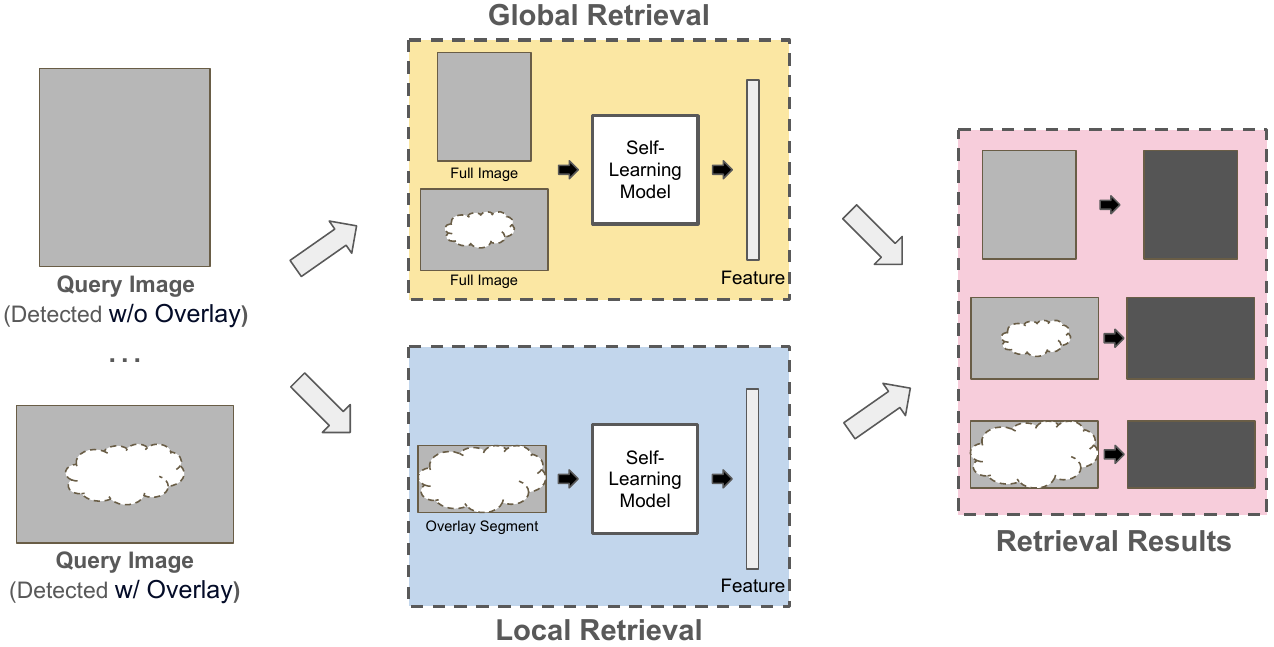}
    \end{center}
    \caption{Fact retrieval is divided into two branches: global retrieval and local retrieval. For an image without any overlays, global retrieval alone can find the original images. For images with one or more overlays, global retrieval is used to search for the entire image, and local retrieval is used to search for the detected forgery segments.}
    \label{fig:factretreval}
\end{figure}

The second phase of our framework is fact retrieval, which involves searching for related images from a reference set (original images database of a closed information retrieval system) on the basis of the outputs from the first phase. As shown in Figure \ref{fig:factretreval}, this phase includes two branches: global retrieval and local retrieval. The results from both branches are combined to form the final results.

\textbf{Global Retrieval}: Global retrieval involves using the entire image as the query to search within the database. For most forgery types, such as object removal and colorization, the whole image is sufficient for retrieval. Recent advancements in self-supervised learning can be leveraged to learn general-purpose features without supervision, making global retrieval well-suited for the large-scale retrieval task.

\textbf{Local Retrieval}: For more complex forgery types like image splicing and copy-move, a forged image may be derived from multiple data sources. Therefore, global retrieval plus additional retrieval of any overlaid objects is more effective than global retrieval alone. In particular, when the overlaid objects are small, relying solely on global retrieval is insufficient. Thus, simultaneous retrieval of the detected forgery segments ensures that all related original images are identified. Specifically, the detected forgery segments are cropped and input into the retrieval model to search for the corresponding original images.

To ensure reproducibility and provide a comprehensive understanding of our work, we have included the \textbf{pseudo code} of our framework in Section I
of the \textbf{Supplementary Materials}, including detailed descriptions of the algorithms used in both phases.
\subsection{Dataset}

\subsubsection{Dataset Sources}

The original images in our dataset were sourced from the Google Open Images Dataset \cite{kuznetsova2020open}. It was selected due to its extensive range of object categories and rich annotations, which are conducive to applying various forgery manipulations.

\begin{figure}[t!]
    \begin{center} 
        \includegraphics[width=0.8\linewidth]{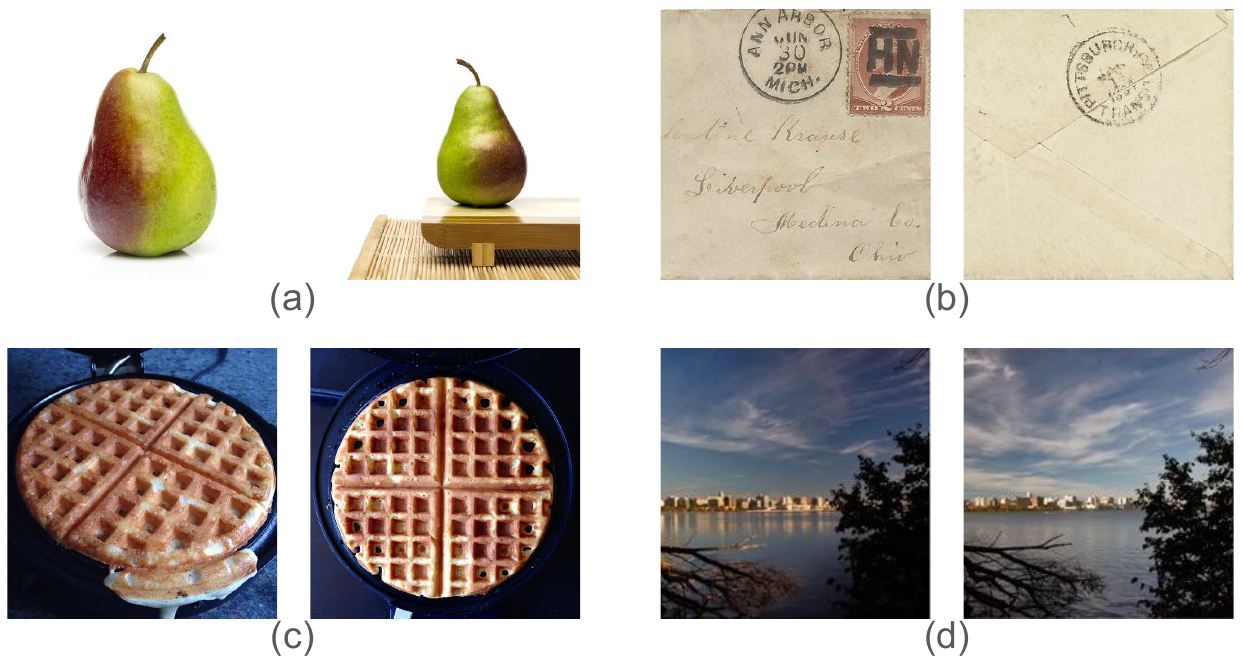}
    \end{center}
    \caption{Examples of similar image pairs that were removed during preprocessing: (a) and (b) show visually similar image pairs, (c) shows images captured from different camera angles, and (d) shows images taken at the same location but at different times.}
    \label{fig:ImageConflicts}
\end{figure}

\textbf{Preprocessing.}: As mentioned in the dataset introduction section of ISC2021, datasets like Google Open Images can contain pairs of similar images, which may lead to conflicts during fact retrieval. Examples of such conflicts are illustrated in Figure \ref{fig:ImageConflicts}. These conflicts arise when two images appear as edited versions of one another but actually represent the same object captured at different times, from different angles, or are simply similar. To address this, we use a pre-trained network to identify and remove the most similar images as a preprocessing step, ensuring the integrity of our dataset for accurate fact verification.

\subsubsection{Structure}
\label{sec:datasetstructure}

To support our two-phase open framework for image-based fact verification, we constructed a comprehensive dataset that mirrors the settings of the ISC2021 competition. The dataset includes several key components, each serving a specific purpose within the framework. There are seven folders containing various source images.
\begin{itemize}
  \item Reference: $800,000$ images without any transformation. Included are source images for queries, selected hard negative samples, and unrelated images that serve as the reference set for fact retrieval tasks.
  \item Training: $800,000$ images similar to those in the reference set. These images are intended for various training tasks, especially for those that depend on the data distribution of reference images, including model training via data augmentation, score normalization, and principal component analysis.
  \item Forgery Images (Query): $40,000$ generated forgery images, with $10,000$ images for each forgery type (copy-move, image splicing, inpainting, and colorization). These images are helpful for both forgery identification and image retrieval.
  \item Augmented Forgery Images: $40,000$ augmented forgery images, manipulated from the original forgery images using a collection of manipulations of varying difficulty levels. These images help enhance the robustness of trained models by simulating real-world conditions.
  \item Original Images: $40,000$ original, unaltered images corresponding to the forgery images. Together with the forgery images, they are used for forgery classification training.
  \item Forgery Segments: $20,000$ supplementary forgery segments from the forgery images, including segments from copy-move and image splicing. They are used for training segment retrieval models.
  \item Annotations: $30,000$ sets of detailed annotations for the forgery images, including masks and bounding boxes, which are essential for training and evaluating forgery localization models.
\end{itemize}

This structured approach ensures that our dataset is versatile and comprehensive, supporting the diverse needs of the proposed multi-task framework.
\subsubsection{Forgery Image Generation}

Our dataset incorporates a diverse array of \textbf{content-aware manipulations} that change the ``content'' of the images to enhance the richness of forgery types. The following describes the processes used to generate four types of forgery images.

\begin{figure}[t]
    \begin{center} 
        \includegraphics[width=1.0\linewidth]{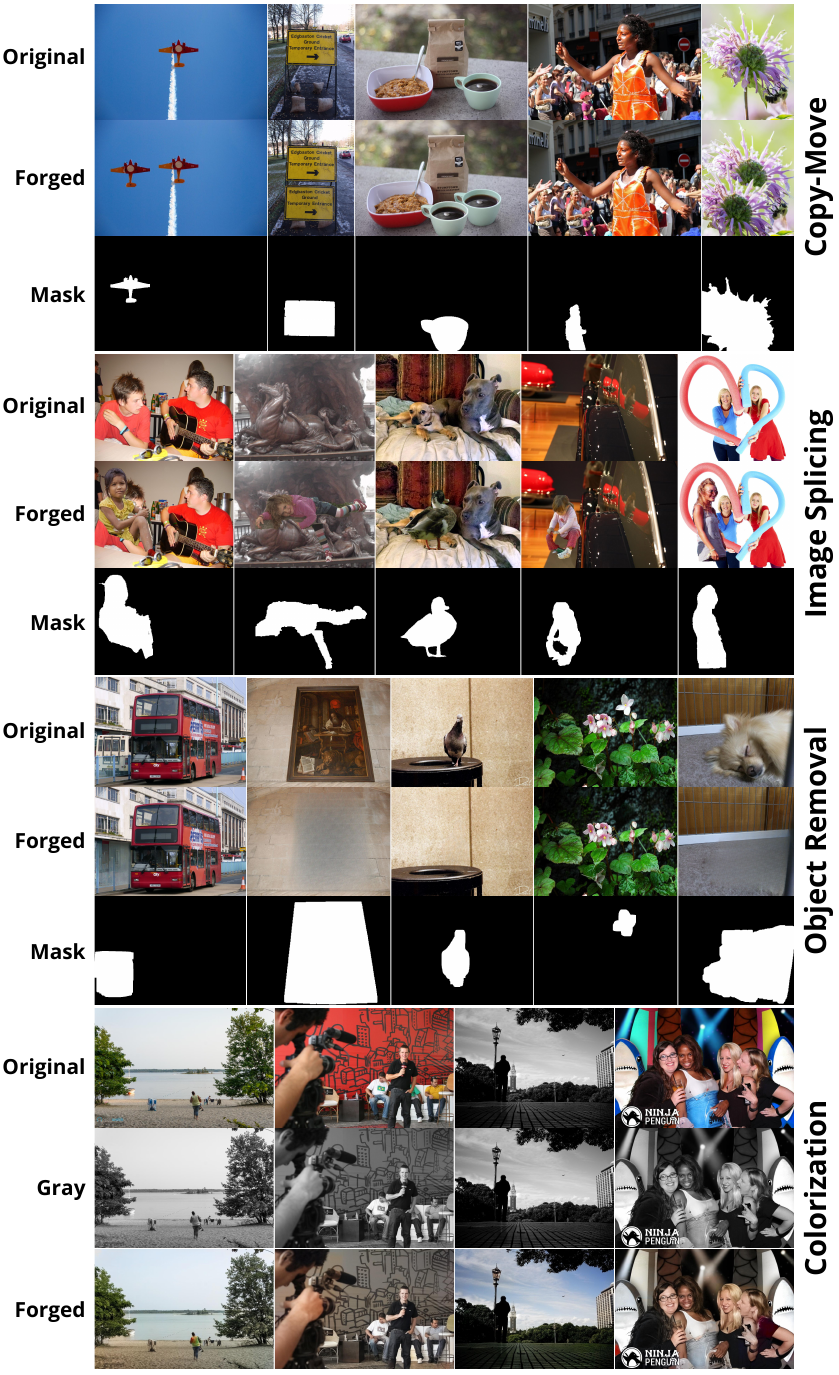}
    \end{center}
    \caption{Examples of four types of forgery images in our dataset. Original image, forged image, and corresponding mask are shown for the first three types. Original color image, grayscale image, and forged image are shown for the last type (colorization).}
    \label{fig:datasetexample}
\end{figure}

\textbf{Copy-Move}: Starting with images that have object masks, we generated copy-moved images \textbf{manually}. To ensure a natural fit, we blended the objects into the background using alpha blending and advanced deep image matting \cite{xu2017deep}. We generated a total of $10,000$ images for this category. Examples are shown in Figure \ref{fig:datasetexample} (rows 1–3).

\textbf{Image Splicing}: Similar to the copy-move process, we \textbf{manually} created
image-splicing forgeries. However, the copied objects were selected from another image within the reference set instead of using ones from the same image. Again, we generate $10,000$ images for this category. Examples are shown in Figure \ref{fig:datasetexample} (rows 4-6).

\textbf{Object Removal}: For object removal, we carefully selected candidate images from specific categories within the reference set, avoiding selections that would result in unnatural outcomes ({\em e.g.}, missing human limbs or clothing). We used two SOTA methods, Lama \cite{suvorov2022resolution} and HiFill \cite{yi2020contextual}, to generate the forgery images, enhancing the diversity of our dataset. We again generated $10,000$ images for this category, with each method contributing $5,000$ images. Examples are shown in Figure \ref{fig:datasetexample} (rows 7-9).

\textbf{Colorization}: For the unique forgery type of colorization, we used two SOTA methods, Colorization Transformer \cite{kumar2021colorization}, and GLEAN \cite{chan2021glean}, and generated $5,000$ forgery images for each one. We converted the original images to grayscale and then applied one of the colorization methods to create artificial colors. Examples are shown in Figure \ref{fig:datasetexample} (rows 10-12).

\subsubsection{External Augmentations for Fact Retrieval}

In real-world scenarios, images on social media often undergo additional manipulations such as compression, cropping, filtering, and text addition. To simulate these conditions and enhance our dataset's robustness, we prepared an augmented dataset for those conditions requiring additional training data. This dataset includes a variety of \textbf{content-preserving} manipulations that do not alter the fundamental ``content'' of the image, thereby reflecting real-world usage more accurately.

To ensure that the augmented images cover a broad spectrum of difficulties, we used manipulations from two toolkits: AugLy \cite{papakipos2022augly} and ImgAug \cite{jung2019imgaug}. These manipulations are categorized into the following groups:
\begin{itemize}
  \item Color processing: Brightness, Saturation, Grayscale, Color Filtering, Contrast Adjustment.
  \item Pixel-level processing: Blur, Compression, Image Coding, Pixelization, Sharpening.
  \item Geometric distortion: Crop, Rotation, Flip, Pad, Aspect Ratio, Perspective.
  \item Image Corruption: Noise Addition, Dropout, Jigsaw Distortion.
  \item Weather Effect: Fog, Rain, Cloudy, Snow.
  \item Embedding the image into the graphical user interface of a social network application.
\end{itemize}

\begin{figure}[t!]
    \begin{center} 
        \includegraphics[width=1.0\linewidth]{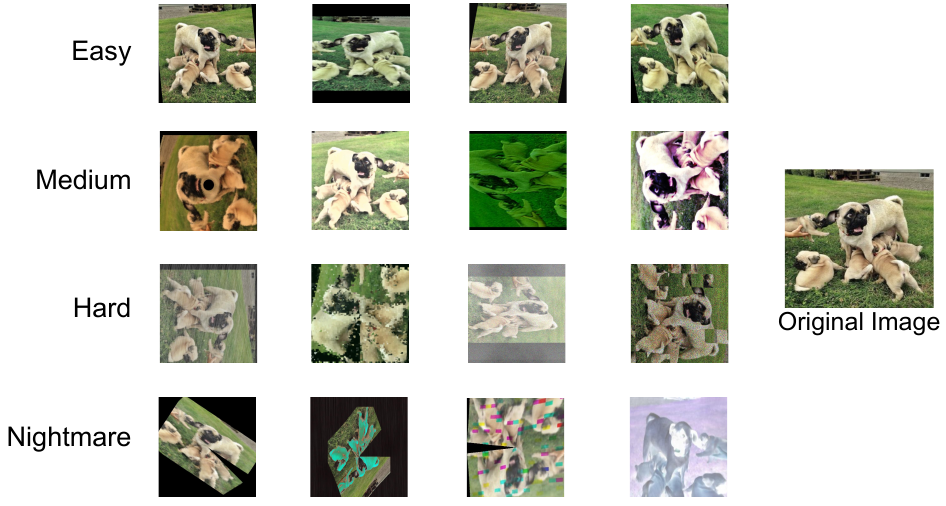}
    \end{center}
    \caption{Examples images illustrating difficulty levels of transformations for data augmentation. Original image is on right. Each row on left shows examples of easy, medium, hard, and nightmare difficulty levels.}
    \label{fig:Augmentation}
\end{figure}

To ensure that the difficulty of the generated augmentations was evenly distributed, we used a pre-trained detection model to examine the extent of changes each augmentation made to the images. Given these results, we categorized the augmentations into four levels: easy, medium, hard, and nightmare. During generation, multiple manipulations were randomly applied within each level to create a realistic and challenging dataset. Figure \ref{fig:Augmentation} shows examples of these augmentations at varying difficulty levels.

\section{Evaluation}
\label{sec:evaluation}
Experiments were conducted to evaluate the performance of our multi-phase framework on the three main tasks: forgery identification, fact retrieval, and fact verification. Within forgery identification, there are two sub-tasks: forgery localization and forgery classification. For each one, we ran several strong baselines or SOTA methods targeting different forgery types against our dataset. Within fact retrieval, there are two sub-tasks: global retrieval and local retrieval. For each one, we ran several baselines from ISC2021 against our dataset. Finally, we compared performance on the fact verification task of a baseline approach that uses only simple retrieval models with that of our multi-phase framework. 

We additionally investigated the effect of the proportion of forgery parts on verification performance. Detailed analysis can be found in Section II
of the \textbf{Supplementary Materials}.

\subsection{Forgery Identification}
The forgery identification phase is aimed at detecting whether an input image is a forgery and, if it is, predicting the forgery type, thereby facilitating the selection of appropriate models for forgery localization and retrieval.

\subsubsection{Forgery Classification}

Forgery classification first involves forgery multiclass classification, followed by binary classification for each forgery type.

\textbf{Data Preparations}: As described in Section \ref{sec:datasetstructure}, the Forgery Images and Original Images folders were used for the classification task. Each forgery type has $10,000$ forgery images and $10,000$ corresponding authentic images.

\textbf{Multiclass Forgery Classification}: We classified images into five categories: authentic, inpainting, image-splicing, copy-move, and colorization. $10,000$ images for each category were split into training, testing, and validation at a ratio of $8:1:1$.
We used two pre-trained models, VGG19 \cite{simonyan2014very} and EfficientNet-B4 \cite{tan2019efficientnet}, with categorical cross-entropy as the loss function and. The Adam optimizer was used with a learning rate of $0.001$. 
The performance metrics were accuracy, precision, recall, and equal error rate (EER), where EER is a common metric used in verification tasks because it provides a balanced measure by considering both the false acceptance rate and false rejection rate simultaneously. Test results are shown in Table \ref{table:classification-first}.

\begin{table}[t]
\caption{Test results for multiclass forgery classification into five categories. Evaluation metrics are accuracy (Acc.), precision (Pre.), and equal error rate (EER). Best performance is highlighted in \textbf{bold}.}
\centering
\label{table:classification-first}
\begin{tabular}{|cc||ccc|ccc|}
\hline
\multicolumn{2}{|c|}{Network}                                                                     & \multicolumn{3}{c|}{VGG19} & \multicolumn{3}{c|}{EfficientNet-B4}     \\ \hline\hline
\multicolumn{1}{|c|}{Real} & \multicolumn{1}{c|}{Forged} & \multicolumn{1}{c|}{Pre.}  & \multicolumn{1}{c|}{Acc.}    & \multicolumn{1}{c|}{EER}    & \multicolumn{1}{c|}{Pre.}  & \multicolumn{1}{c|}{Acc.}    & \multicolumn{1}{c|}{EER}   \\ \hline
\multicolumn{1}{|c|}{2.5k}    & \multicolumn{1}{c|}{10k}  & \multicolumn{1}{c|}{81.26} & \multicolumn{1}{c|}{81.12} & \multicolumn{1}{c|}{27.30} & \multicolumn{1}{c|}{94.42} & \multicolumn{1}{c|}{92.56} & \multicolumn{1}{c|}{9.60}  \\ \hline
\multicolumn{1}{|c|}{5k}    & \multicolumn{1}{c|}{20k} & \multicolumn{1}{c|}{83.19} & \multicolumn{1}{c|}{83.16} & \multicolumn{1}{c|}{27.90} & \multicolumn{1}{c|}{97.27} & \multicolumn{1}{c|}{94.88} & \multicolumn{1}{c|}{6.35} \\ \hline
\multicolumn{1}{|c|}{7.5k}    & \multicolumn{1}{c|}{30k} & \multicolumn{1}{c|}{92.98} & \multicolumn{1}{c|}{90.59} & \multicolumn{1}{c|}{12.67} & \multicolumn{1}{c|}{97.13} & \multicolumn{1}{c|}{95.36} & \multicolumn{1}{c|}{4.93} \\ \hline
\multicolumn{1}{|c|}{10k}    & \multicolumn{1}{c|}{40k} & \multicolumn{1}{c|}{\textbf{94.25}} & \multicolumn{1}{c|}{\textbf{91.72}} & \multicolumn{1}{c|}{\textbf{11.70}} & \multicolumn{1}{c|}{\textbf{97.88}}    & \multicolumn{1}{c|}{\textbf{95.96}} & \multicolumn{1}{c|}{\textbf{4.35}} \\ \hline
\end{tabular}
\end{table}

\begin{table}[t!]
\caption{Test results for four binary forgery classification experiments conducted for each forgery type. Evaluation metrics are accuracy, precision, and EER. Best performance is highlighted in \textbf{bold}.}
\centering
\label{table:classification-second}
\begin{tabular}{|c|ccc|ccc|}
\hline
\multirow{2}{*}{Network} & \multicolumn{3}{c|}{VGG19}                                           & \multicolumn{3}{c|}{EfficientNet-B4}                                         \\ \cline{2-7} 
                              & \multicolumn{1}{c|}{Pre.} & \multicolumn{1}{c|}{Acc.} & EER   & \multicolumn{1}{c|}{Pre.} & \multicolumn{1}{c|}{Acc.} & EER   \\ \hline\hline
Copy-Move            & \multicolumn{1}{c|}{65.57}     & \multicolumn{1}{c|}{72.08}    & 24.46 & \multicolumn{1}{c|}{96.60}     & \multicolumn{1}{c|}{94.80}    & 4.85 \\ \hline
Image Splicing       & \multicolumn{1}{c|}{84.20}     & \multicolumn{1}{c|}{85.81}    & 14.14 & \multicolumn{1}{c|}{97.55}     & \multicolumn{1}{c|}{96.74}    & 3.41 \\ \hline
Object Removal           & \multicolumn{1}{c|}{88.37}     & \multicolumn{1}{c|}{82.94}    & 18.74 & \multicolumn{1}{c|}{98.42}     & \multicolumn{1}{c|}{93.69}    & 6.21 \\ \hline
Colorization         & \multicolumn{1}{c|}{\textbf{96.53}}     & \multicolumn{1}{c|}{\textbf{91.15}}    & \textbf{11.37} & \multicolumn{1}{c|}{\textbf{98.86}}     & \multicolumn{1}{c|}{96.83}    & 3.56 \\ \hline
ALL Images           & \multicolumn{1}{c|}{74.95}     & \multicolumn{1}{c|}{76.89}    & 23.28 & \multicolumn{1}{c|}{96.80}     & \multicolumn{1}{c|}{\textbf{96.84}}    & \textbf{3.15} \\ \hline
\end{tabular}
\end{table}

\textbf{Binary Forgery Classification}: We conducted binary classification for each forgery type. The setting for the multiclass classification task was used, except that cross-entropy was used as the loss function. The test results are shown in Table \ref{table:classification-second}.

\textbf{Observations}: The results in Table \ref{table:classification-first} indicate that both pre-trained models can effectively predict forgery types, with EfficientNet-B4 achieving significantly higher performance (over $90$\% accuracy). Increasing the number of training samples further enhanced the test outcomes. Similarly, the binary classification results in Table \ref{table:classification-second} are high and even higher than the multiclass one, possibly due to the smaller number of targets. Notably, copy-move detection is the most challenging type, likely due to the high fidelity of manually generated copy-move instances.

\subsubsection{Forgery Localization}

Given the distinct nature of each forgery type, individual evaluation was conducted. SOTA methods were used to evaluate each forgery type by comparing our dataset with existing datasets. Specifically, we evaluated copy-move, image splicing, and object removal (excluding colorization, which lacks a detection target).

\begin{table}[t]
\caption{Quantitative comparisons using precision, recall, and F1-score for copy-move forgery localization. The highest value for each column is highlighted in \textbf{bold}.}
\label{table:copy-move-quantitative}
\centering
\begin{tabular}{ccccc}
\hline
\multirow{2}{*}{Models} & \multirow{2}{*}{Metric} & \multicolumn{3}{c}{Test Dataset} \\ 
\cline{3-5} 
& & CASIA  &  CoMoFoD  & Ours    \\ 
\hline
BusterNet \cite{wu2018busternet}            & Precision                                    & \textbf{55.71}    & \textbf{51.25}     &  39.12       \\
DOA-GAN \cite{islam2020doa}                 & Precision                                    & 54.70    & 48.42     &  \textbf{41.94}       \\
Serial Network \cite{chen2020serial}        & Precision                                    & 53.08    & 46.10     &  36.41       \\ \hline
BusterNet \cite{wu2018busternet}            & Recall                                       & 43.83    & 41.67     &  32.56       \\
DOA-GAN \cite{islam2020doa}                 & Recall                                       & 39.67    & 37.84     &  29.96       \\
Serial Network \cite{chen2020serial}        & Recall                                       & \textbf{49.79}    & \textbf{42.20}     &  \textbf{38.12}       \\ \hline
BusterNet                                   & F1-score                                           & 45.56    & 43.78     &  \textbf{34.52}       \\
DOA-GAN                                     & F1-score                                           & 41.44    & 36.92     &  21.45       \\
Serial Network                              & F1-score                                           & \textbf{47.68}    & \textbf{44.10}     &  35.32       \\ \hline
\end{tabular}
\end{table}

\textbf{Copy-Move Forgery Localization}: Three SOTA deep learning frameworks, (BusterNet \cite{wu2018busternet}, DOA-GAN \cite{islam2020doa}, and Serial Network \cite{chen2020serial}) were used for copy-move detection. Using the codes provided by the authors, we trained these models on the ``copy-move'' split of our dataset, comprising $10,000$ forgery instance pairs. We compared the performance of our dataset with those of the CASIA \cite{dong2013casia} and CoMoFoD \cite{tralic2013comofod} datasets, rescaling all images for a fair evaluation.

\begin{figure*}[t]
    \begin{center} 
        \includegraphics[width=0.9\linewidth]{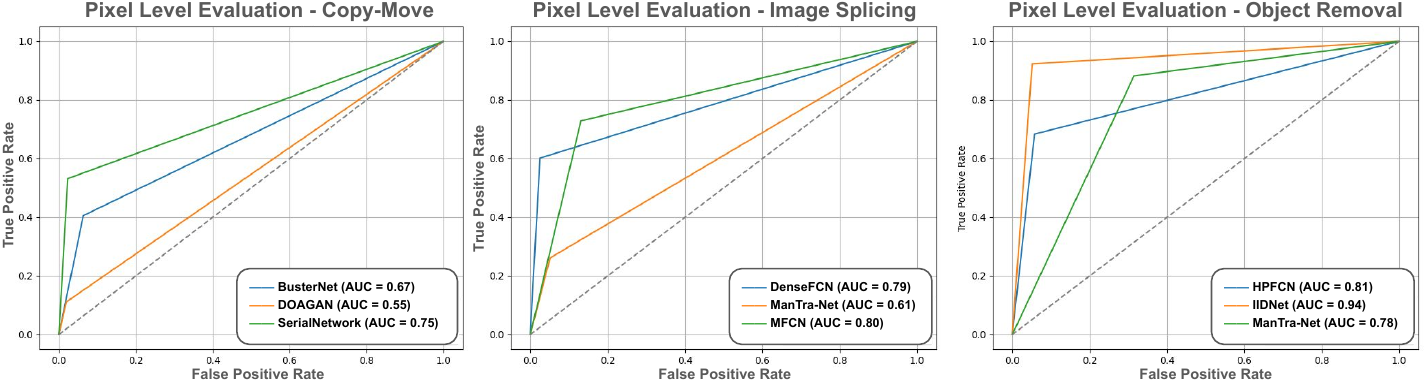}
    \end{center}
    \caption{Comparison of pixel-level AUC performance on our dataset. From left to right: copy-move, image splicing, and object removal experiment results.}
    \label{fig:PRCurve}
\end{figure*}

\begin{figure}[t!]
    \begin{center} 
        \includegraphics[width=1.0\linewidth]{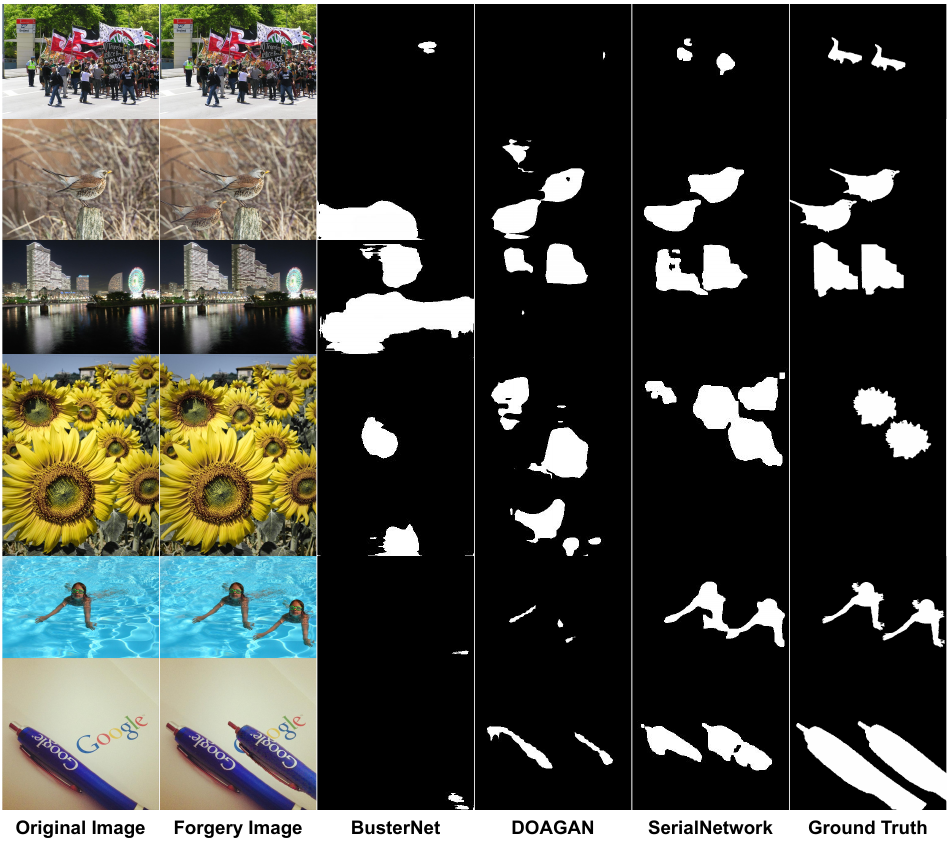}
    \end{center}
    \caption{Visualization examples from our dataset, showing original images, forgery images, predicted maps from BusterNet \cite{wu2018busternet}, DOA-GAN \cite{islam2020doa}, and Serial Network \cite{chen2020serial}, and ground truth (from left to right). The forgery images here were generated by copying a region of the image and pasting it into another part of the same image.}
    \label{fig:copy-move-qualitative}
\end{figure}

For quantitative analysis, we used precision, recall, and the F1-score as the evaluation metrics. The higher the score, the better the performance.
As shown in Table \ref{table:copy-move-quantitative}, upon shifting to our dataset, all SOTA methods experienced a significant performance drop for all three metrics. Specifically, compared with those for the CASIA \cite{dong2013casia} and CoMoFoD \cite{tralic2013comofod} datasets, the average F1-scores for our dataset were $10.15$\%, $17.73$\%, and $10.57$\% lower than those for BusterNet \cite{wu2018busternet}, DOA-GAN \cite{islam2020doa}, and Serial Network\cite{chen2020serial}, respectively. Figure \ref{fig:PRCurve} shows that AUC performance at the pixel level was low for all methods on our dataset. Figure \ref{fig:copy-move-qualitative} shows map predictions generated by BusterNet \cite{wu2018busternet}, DOA-GAN \cite{islam2020doa}, and Serial Network \cite{chen2020serial}, demonstrating the difficulty of attaining accurate predictions with our dataset.

\begin{table}[]
\caption{Quantitative comparisons using IoU and F1-scores for image splicing forgery localization. The highest value in each column is highlighted in \textbf{bold}.}
\label{table:image-splicing-quantitative}
\centering
\begin{tabular}{cccccc}
\hline
\multirow{2}{*}{Models} & \multirow{2}{*}{Metric} & \multicolumn{4}{c}{Test Dataset}              \\ \cline{3-6} 
                        &                          & CASIA1 & Carvalho & Columbia & Ours  \\ \hline
MFCN \cite{salloum2018image}     & IOU                      & \textbf{30.43}  & \textbf{21.50}    & \textbf{52.90}    & 50.67 \\
Mantra-Net \cite{wu2019mantra}   & IOU                      & 12.61  & 20.17    & 32.80    & 38.12 \\
DenseFCN \cite{zhuang2021image}  & IOU                      & 6.88   & 16.38    & 23.12    & \textbf{61.17}\\ \hline
MFCN \cite{salloum2018image}     & F1-score                 & \textbf{37.55}  & 32.11    & \textbf{62.22}    & 63.69 \\
Mantra-Net \cite{wu2019mantra}   & F1-score                 & 20.09  & \textbf{32.51}    & 47.18    & 26.67 \\
DenseFCN \cite{zhuang2021image}  & F1-score                 & 9.83   & 27.11    & 32.29    & \textbf{73.48} \\ \hline
\end{tabular}
\end{table}

\textbf{Image Splicing Forgery Localization}: Three SOTA deep learning frameworks (MFCN \cite{salloum2018image}, Mantra-Net \cite{wu2019mantra}, and DenseFCN \cite{zhuang2021image}) were used for image splicing detection. Using the authors' codes, we trained these models on the ``image splicing'' split of our dataset, comprising $10,000$ forgery instance pairs. We compared the performance of our dataset with those of the CASIA v1.0 \cite{dong2013casia}, Carvalho \cite{de2013exposing}, and Columbia \cite{hsu2006detecting} datasets, rescaling all images for a fair evaluation.

\begin{figure}
    \begin{center} 
        \includegraphics[width=1.0\linewidth]{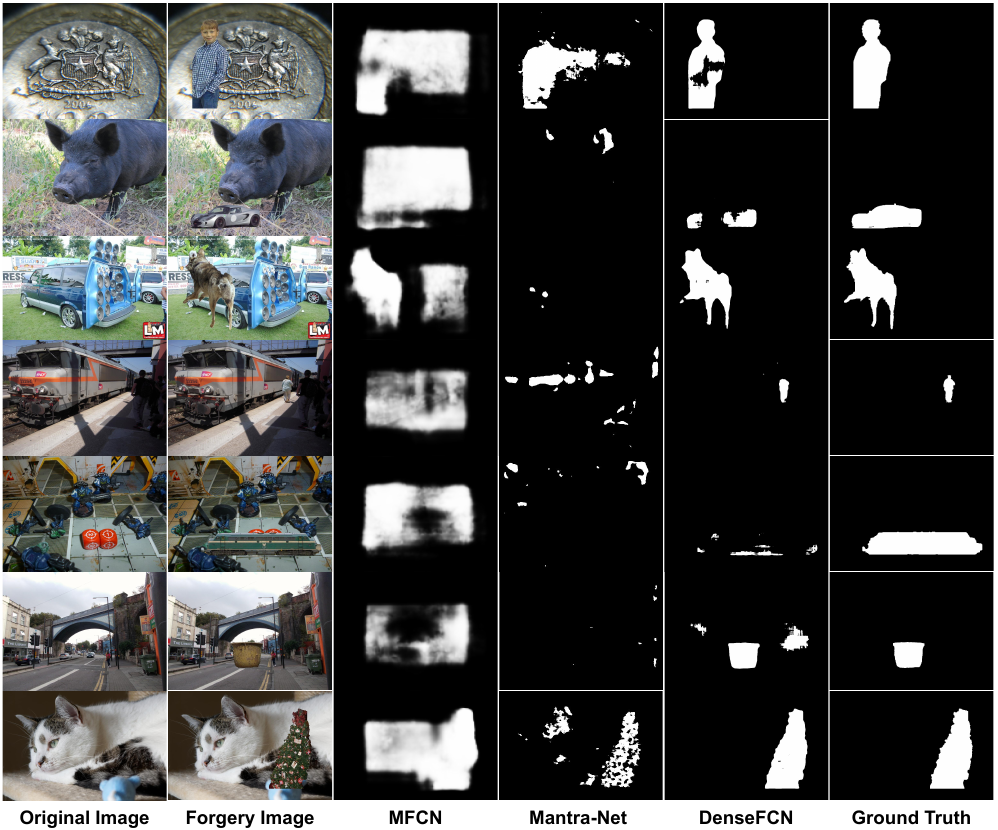}
    \end{center}
    \caption{Visualization examples from our dataset showing original images, forgery images, predicted maps from MFCN \cite{salloum2018image}, Mantra-Net \cite{wu2019mantra}, DenseFCN \cite{zhuang2021image}, and ground truth (from left to right). The forgery images here were generated by cutting parts from different images and pasting them into a single image.}
    \label{fig:image-splicing-qualitative}
\end{figure}

For quantitative analysis, we used the IoU and F1-score as the evaluation metrics. The higher the score, the better the performance.
The results are shown in Table \ref{table:image-splicing-quantitative}. Figure \ref{fig:PRCurve} shows that AUC performance at the pixel level was low across all methods on our dataset. Figure \ref{fig:image-splicing-qualitative} shows map predictions generated by MFCN \cite{salloum2018image}, Mantra-Net \cite{wu2019mantra}, and DenseFCN \cite{zhuang2021image}, demonstrating the difficulty of attaining accurate predictions with our dataset.

\textbf{Object Removal Forgery Localization}: Three SOTA deep learning frameworks (MT-Net \cite{wu2019mantra}, HP-FCN \cite{li2019localization}, and IID-Net \cite{wu2021iid}) were used for object removal detection. Using the codes provided by the authors, we trained these models on the ``object removal'' split of our dataset, comprising $10,000$ forgery instance pairs. We compared the performance of our dataset with those of the GC \cite{yu2019free}, SH \cite{yan2018shift}, and LB \cite{learningdeep} datasets, rescaling all images for a fair evaluation.

\begin{figure}[t]
    \begin{center} 
        \includegraphics[width=1\linewidth]{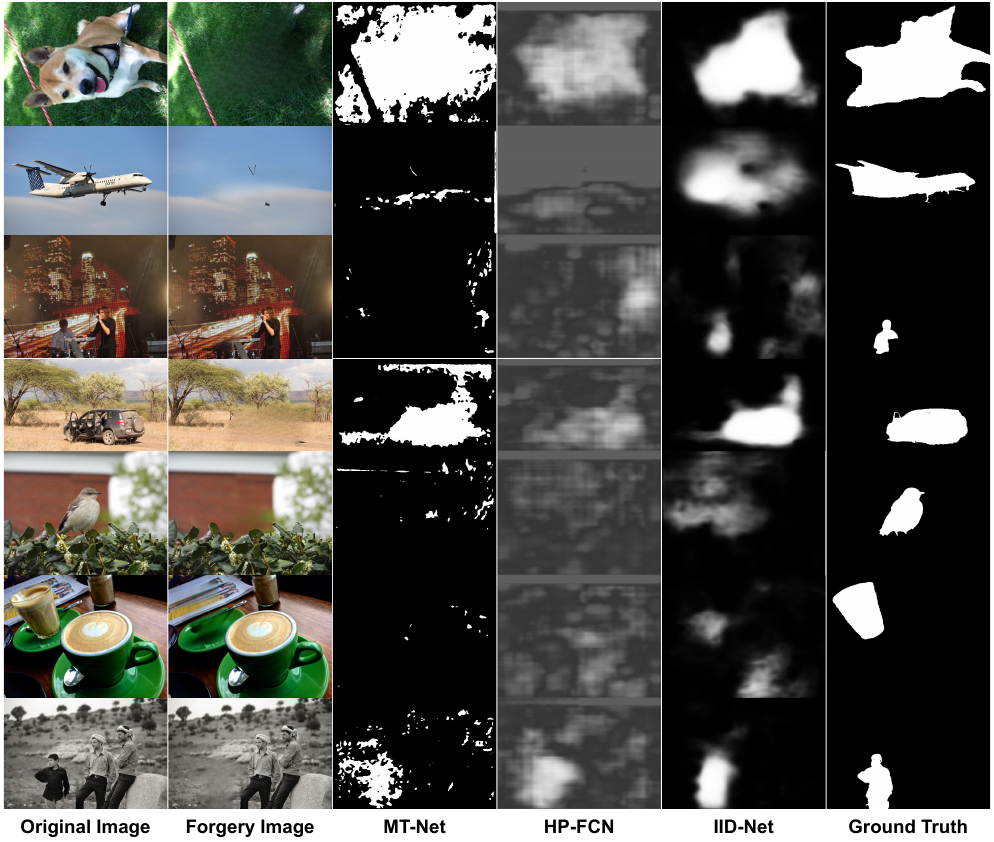}
    \end{center}
    \caption{Visualization examples from our dataset showing original images, forgery images, predicted maps from MT-Net \cite{wu2019mantra}, HP-FCN \cite{li2019localization}, IID-Net \cite{wu2021iid}, and ground truth (from left to right). The forgery images here were generated by removing objects from the image and filling in the area with surrounding pixels.}
    \label{fig:image-inpainting-qualitative}
\end{figure}

For quantitative analysis, we used AUC and the F1-score as the evaluation metrics. The higher the score, the better the performance.
The results are shown in Table \ref{table:image-inpainting-quantitative}. Upon shifting from most datasets to our dataset, basically all SOTA methods experienced a drop for the two metrics. Specifically, compared with those for the GC \cite{yu2019free}, SH \cite{yan2018shift}, and LB \cite{learningdeep} datasets, the average F1-score for our dataset was $0.545$\%, $7.73$\%, and $26.03$\% lower MT-Net \cite{wu2019mantra}, HP-FCN \cite{li2019localization}, and IID-Net \cite{wu2021iid}, respectively. 
We also observed that for MT-Net, LB achieved a slightly lower AUC and F1-score than our dataset, which exhibited an obvious average performance drop for all cases.
Figure \ref{fig:PRCurve} shows that AUC performance at the pixel level was low across all methods on our dataset. Figure \ref{fig:image-inpainting-qualitative} shows map predictions generated by MT-Net \cite{wu2019mantra}, HP-FCN \cite{li2019localization}, and IID-Net \cite{wu2021iid}, demonstrating the difficulty of attaining accurate predictions with our dataset.

\begin{table}[]
\caption{Quantitative comparisons using AUC and F1-score for object removal forgery localization. The highest value in each column is highlighted in \textbf{bold}.}
\label{table:image-inpainting-quantitative}
\centering
\begin{tabular}{cccccc}
\hline
\multirow{2}{*}{Models} & \multirow{2}{*}{Metric} & \multicolumn{4}{c}{Test Dataset} \\ \cline{3-6} 
                        &                          & GC     & SH     & LB     & Ours  \\ \hline
MT-Net \cite{wu2019mantra}         & AUC                      & 96.31  & 73.58  & 62.27  & 65.75 \\
HP-FCN \cite{li2019localization}   & AUC                      & 96.65  & 98.14  & 96.51  & \textbf{71.73} \\
IID-Net \cite{wu2021iid}           & AUC                      & \textbf{96.77}  & \textbf{99.67}  & \textbf{99.80}  & 62.06 \\ \hline
MT-Net \cite{wu2019mantra}         & F1-score                 & 14.17  & 72.63  & 60.14  & 65.84 \\
HP-FCN \cite{li2019localization}   & F1-score                 & 76.93  & 81.43  & 55.78  & 63.65 \\
IID-Net \cite{wu2021iid}           & F1-score                 & \textbf{83.61}  & \textbf{94.13}  & \textbf{96.14}  & \textbf{65.26} \\ \hline
\end{tabular}
\end{table}

\subsection{Fact Retrieval}
The aim of the fact retrieval phase is to retrieve the original images from the reference set on the basis of the outputs of the forgery identification phase. The first experiment tested global (entire-image) retrieval of the entire image from the reference set. The second one tested local (segment) retrieval of specific image segments from the reference set.

\subsubsection{Evaluation Metrics}
In the fact retrieval phase, the goal is to find all relevant images, similar to the goal in ICD. This means locating the source image rather than simply a visually similar one. Therefore, we used the micro-average precision (µAP) metric introduced in ISC2021 \cite{douze20212021}.

\textbf{micro-Average Precision (µAP)}: The output of ICD includes a list of sets, each consisting of a query image, a chosen candidate source image, and a confidence value. Not all queries are included in this list due to distractors. µAP is used as it comprehensively assesses the detection system's performance, focusing on confidence values more than the ranking per query, unlike mean average precision (mAP). 

\subsubsection{Baseline Methods}
We selected two baseline methods in accordance with the decisions made in ISC2021.

\textbf{GIST}: The GIST descriptor \cite{oliva2001modeling}, which stands for ``Global Image Structure Tensor," is an economical and straightforward descriptor suitable for low-resolution images in copy detection due to its cost-effective extraction and freedom from training. 

\textbf{MultiGrain}: The MultiGrain descriptor \cite{berman2019multigrain} is an early example of global image embedding methods that heavily rely on data augmentation. It uses a ResNet50 model pre-trained on ImageNet \cite{deng2009imagenet}. This model includes an additional head for generalized max-pooling on the final activation map to generate an image embedding.

\subsubsection{Performance of Global Retrieval}

\begin{table*}
\centering
\caption{Global retrieval performance in terms of µAP on copy-move and image splicing for all queries from our dataset and all queries from ISC2021 using GIST and MultiGrain as baselines.}
\label{table:fullretrieval}
\begin{tabular}{|c|cccc|cccc|}
\hline
Method            & \multicolumn{4}{c|}{GIST}   & \multicolumn{4}{c|}{MultiGrain}                                                                                                               \\ \hline
Forgery Type      & \multicolumn{1}{c|}{Copy-Move} & \multicolumn{1}{c|}{Image Splicing} & \multicolumn{1}{c|}{Ours} & ISC2021 & \multicolumn{1}{c|}{Copy-Move} & \multicolumn{1}{c|}{Image Splicing} & \multicolumn{1}{c|}{Ours} & ISC2021 \\ \hline
Average Precision & \multicolumn{1}{c|}{0.00137}            & \multicolumn{1}{c|}{0.26800}                 & \multicolumn{1}{c|}{0.47760}                           & \textbf{0.59836} & \multicolumn{1}{c|}{0.25640}            & \multicolumn{1}{c|}{0.47360}                 & \multicolumn{1}{c|}{0.70243}                           & \textbf{0.81952}             \\ \hline
Recall at P90     & \multicolumn{1}{c|}{0.00081}            & \multicolumn{1}{c|}{0.22988}                 & \multicolumn{1}{c|}{0.38266}                           & \textbf{0.48132}  & \multicolumn{1}{c|}{0.00725}            & \multicolumn{1}{c|}{0.30777}                 & \multicolumn{1}{c|}{0.57624}                           & \textbf{0.62951}             \\ \hline
Threshold at P90  & \multicolumn{1}{c|}{-0.0564344}         & \multicolumn{1}{c|}{-0.105304}               & \multicolumn{1}{c|}{-0.0794474}                        & \textbf{-0.1513141} & \multicolumn{1}{c|}{-0.644967}          & \multicolumn{1}{c|}{-1.07013}                & \multicolumn{1}{c|}{-0.874921}                         & \textbf{-1.10512}            \\ \hline
Recall at rank 1  & \multicolumn{1}{c|}{0.01289}            & \multicolumn{1}{c|}{0.33207}                 & \multicolumn{1}{c|}{0.57994}                           & \textbf{0.68203}  & \multicolumn{1}{c|}{0.79533}            & \multicolumn{1}{c|}{0.72573}                 & \multicolumn{1}{c|}{0.83652}                           & \textbf{0.85194}             \\ \hline
Recall at rank 10 & \multicolumn{1}{c|}{0.01853}            & \multicolumn{1}{c|}{0.35116}                 & \multicolumn{1}{c|}{0.59004}                           & \textbf{0.64485}  & \multicolumn{1}{c|}{0.84045}            & \multicolumn{1}{c|}{0.76832}                 & \multicolumn{1}{c|}{0.86341}                           & \textbf{0.87956}             \\ \hline
\end{tabular}
\end{table*}

We applied the two baseline methods to both the ``query'' split of our dataset and the ``Track 1'' split of the ISC2021 dataset, focusing on two forgery types: copy-move and image-splicing.

As shown in Table \ref{table:fullretrieval}, with the same settings and dataset sizes, the performances of both baselines on our dataset were lower than those reported in ISC2021, underscoring the difficulties inherent in our dataset that make retrieval more challenging. MultiGrain showed much better performance than GIST. The individual results for copy-move and image splicing were lower than those for all queries combined, indicating that these forgery types pose significant challenges for global retrieval due to substantial occlusion. Additionally, copy-move scored lower than image splicing, indicating that copy-move is more challenging to handle for global retrieval.

\subsubsection{Performance of Local Retrieval}

For the evaluation of local retrieval, we applied the two baselines to the ``segment'' split of our dataset. Since there is no corresponding part in the ISC2021 dataset, we only evaluated performance on our dataset for two forgery types: copy-move and image-splicing (colorization and object removal have no forgery parts worth retrieving).

As shown in Table \ref{table:segmentretrieval}, the results indicate that both methods demonstrated better performance on segment retrieval than on global retrieval, possibly due to the absence of occlusions. MultiGrain performed much better than GIST, reaching $90$\% when retrieving segments for image spicing. However, image splicing scored lower than copy-move, indicating that image splicing is more challenging to handle for local retrieval.

\begin{table}[t]
\centering
\caption{Local retrieval performance in terms of µAP on copy-move and image splicing using GIST and MultiGrain as baselines.}
\label{table:segmentretrieval}
\begin{tabular}{|c|cc|cc|}
\hline
Method            & \multicolumn{2}{c|}{GIST}                                 & \multicolumn{2}{c|}{MultiGrain}                          \\ \hline
Forgery Type      & \multicolumn{1}{c|}{Copy-Move}  & Splicing & \multicolumn{1}{c|}{Copy-Move} & Splicing \\ \hline
Average Pre. & \multicolumn{1}{c|}{\textbf{0.47440}}    & 0.40865                 & \multicolumn{1}{c|}{\textbf{0.95722}}   & 0.63317                 \\ \hline
Rec. at P90     & \multicolumn{1}{c|}{\textbf{0.33602}}    & 0.29851                 & \multicolumn{1}{c|}{\textbf{0.90330}}   & 0.36386                 \\ \hline
Thr. at P90  & \multicolumn{1}{c|}{\textbf{-0.0862348}} & -0.0885148              & \multicolumn{1}{c|}{\textbf{-0.850185}} & -0.869909               \\ \hline
Rec. rank 1  & \multicolumn{1}{c|}{\textbf{0.63940}}    & 0.56071                 & \multicolumn{1}{c|}{\textbf{0.99919}}   & 0.81767                 \\ \hline
Rec. rank 10 & \multicolumn{1}{c|}{\textbf{0.66559}}    & 0.58975                 & \multicolumn{1}{c|}{\textbf{1.00000}}   & 0.95119                 \\ \hline
\end{tabular}
\end{table}
\subsection{Fact Verification}

Our final experiment focused on our overall goal: fact verification. We compared the performance of our two-phase framework with that of a baseline framework.

\subsubsection{Experiment Preparation}

To facilitate comparisons, we used the ``Train,'' ``Query,'' and ``Reference'' folders ($40,000$, $40,000$, and $800,000$ images, respectively) of our dataset for training and testing. Additionally, we used a ground truth table showing the correspondence between the forgery queries and original images. Each query may correspond to one or two original images in the reference set, or there may be no corresponding original images. We again used µAP as the evaluation metric. Two frameworks were used for comparison.

\textbf{Our Two-Phase Framework}: In the first phase, the authenticity of the input image is predicted. If it is real, nothing is returned; if it is fake, its forgery type is predicted. For copy-move or image splicing forgeries, the corresponding network is used to predict the forgery segments. Finally, the original image(s) of the forgery one is retrieved from the reference set, including the original images of detected segments, if available.

\textbf{Baseline Retrieval Framework}: GIST and MultiGrain were used as simple baselines in this framework to search for original images in the reference set.

\subsubsection{Performance of Fact Verification}

As shown in Table \ref{table:factverification}, our two-phase framework showed much better performance than the baseline one. This is because the baseline retrieval framework can retrieve at most one original image from the reference set. This limitation makes it difficult to search for all the original images, especially for forgery types like copy-move and image splicing, resulting in performance equivalent to full-image retrieval. Our two-phase framework performed better because it can search all related original images. This flexibility can be adjusted to accommodate more complex and diverse forgery images and can be reset for open-set retrieval in the future. 

\begin{table}
\centering
\caption{Fact verification performance in terms of µAP on our dataset using the two-phase framework and baseline retrieval framework.}
\label{table:factverification}
\begin{tabular}{|c|cc|cc|}
\hline
Method            & \multicolumn{2}{c|}{GIST}                                 & \multicolumn{2}{c|}{MultiGrain}                          \\ \hline
Framework      & \multicolumn{1}{c|}{Two-phase}  & Baseline & \multicolumn{1}{c|}{Two-phase} & Baseline \\ \hline
Ave Pre. & \multicolumn{1}{c|}{\textbf{0.62351}}    & 0.47760                 & \multicolumn{1}{c|}{\textbf{0.82531}}   & 0.70243                 \\ \hline
Rec. at P90     & \multicolumn{1}{c|}{\textbf{0.59582}}    & 0.38266                 & \multicolumn{1}{c|}{\textbf{0.72295}}   & 0.57624                 \\ \hline
Thr. at P90  & \multicolumn{1}{c|}{\textbf{-0.0462482}} & -0.0794474              & \multicolumn{1}{c|}{\textbf{-0.819524}} & -0.874921               \\ \hline
Rec. rank 1  & \multicolumn{1}{c|}{\textbf{0.75424}}    & 0.57994                 & \multicolumn{1}{c|}{\textbf{0.89942}}   & 0.83652                 \\ \hline
Rec. rank 10 & \multicolumn{1}{c|}{\textbf{0.79475}}    & 0.59004                 & \multicolumn{1}{c|}{\textbf{0.92932}}   & 0.86341                 \\ \hline
\end{tabular}
\end{table}

\section{Discussion}
\label{sec:discussion}
Forgery techniques are constantly evolving, necessitating the development of robust and adaptable fact verification systems. Our open framework provides a foundational approach to navigating the complexities of forgery identification and fact retrieval. Moving forward, there are several avenues for enhancing our framework.

\begin{enumerate}
    \item \textbf{Sustained Model Improvement}: We can continually update and iterate upon internal networks for forgery identification and fact retrieval. This involves introducing new detection models for emerging forgery types and reinforcing existing networks to handle challenging forgery scenarios. Additionally, comparing differences between the query and retrieved original images can validate and enhance forgery detection models.
    \item \textbf{User Involvement}: The intermediate processing stage between the two phases in our proposed framework offers opportunities for user involvement. We can create a more responsive and user-centric workflow by enabling users to refine their queries iteratively on the basis of the results generated in the intermediate stage. This interactive approach promotes efficiency and enables users to navigate complex forgery scenarios with confidence.
\end{enumerate}

By continuously refining our models and incorporating user feedback in the intermediate stage, we aim to create a more dynamic and effective system that can keep pace with the evolving landscape of forgery techniques.

\section{Conclusion}
\label{sec:conclusion}
The swift progression of forgery technology, fueled by advancements in deepfake generation and classic forgery techniques aided by deep learning, has sparked substantial concerns surrounding the proliferation of deceptive and malicious content. To tackle this challenge, we have introduced the concept of image-based fact verification, with the goal of not only identifying forgeries but also retrieving original images to bolster the credibility of detection models.

Our two-phase open framework seamlessly integrates existing forgery detection networks with a retrieval system inspired by insights gleaned from the Image Similarity Challenge 2021 (ISC2021). While ISC2021 primarily focused on image traceability, our framework extends this notion to authenticating image veracity. Additionally, we have created a novel dataset encompassing a diverse array of forgery types and difficulty levels, serving as an excellent resource for evaluating forgery detection and fact retrieval models. It has proven to be highly effective in supporting and advancing related research endeavors.

Looking ahead, the future evolution of image-based fact verification stands to benefit from ongoing advancements in forgery detection and retrieval techniques. Possibilities include the exploration of more sophisticated forgery detection and retrieval models, expansion of our dataset to encompass a broader spectrum of forgery types, and integration of multi-modal information. These endeavors are essential in mitigating the detrimental effects of forgery technology misuse and in reinforcing trust in the rapidly advancing field of AI.

\appendices

\bibliography{bibliography}
\bibliographystyle{IEEEtran}

\end{document}


\title{Supplementary Materials}

\markboth{IEEE TRANSACTIONS ON INFORMATION FORENSICS AND SECURITY,~Vol.~, No.~, ~}%
{CUI \MakeLowercase{\textit{et al.}}: LookupForensics: A Large-Scale Multi-Task Dataset for Multi-Phase Image-Based Fact Verification}

\maketitle

\section{Algorithms for Our Image-based Automated Fact Verification Framework}
\label{sec:algorithm}

In this Supplement, we provide a detailed description of the algorithms used for image-based automated fact verification, enabling readers to understand and reproduce our results. The proposed method consists of two main phases: forgery identification and fact retrieval. Each phase is designed to process an input image, determine if it is a forgery, and, if so, retrieve the original authentic image(s) efficiently.

\subsection{Algorithm Overview}

The algorithm is primarily aimed at determining the authenticity of an input image and, if the image is forged, retrieving the original images from which it may have been derived. The process is divided into two phases, supported by a series of specialized algorithms:
\begin{enumerate}
    \item \textbf{Forgery Identification}: Detect whether the image is forged and, if so, identify the type and location of the forgery. This phase is divided into three key steps:
    \begin{enumerate}
        \item \textbf{Binary Classification}: Determines if the image is authentic or forged.
        \item \textbf{Forgery Type Classification}: Identifies the type of forgery (e.g., copy-move, image splicing).
        \item \textbf{Forgery Segmentation}: Locates and segments the tampered regions in the image.
    \end{enumerate}
    \item \textbf{Fact Retrieval}: Retrieve the original images using the identified forgery type and location.
    \begin{enumerate}
        \item \textbf{Global Retrieval}: Retrieves potential original images for the entire forged image.
        \item \textbf{Local Retrieval}: Retrieves original images for specific tampered segments identified in the forgery segmentation step.
    \end{enumerate}
\end{enumerate}
In total, our framework employs 15 algorithms to comprehensively handle these phases, ensuring robust and accurate detection and verification of image forgeries. The detailed steps and functionalities of these algorithms are outlined in the subsequent sections.

\subsection{Detailed Algorithm Steps}
\subsubsection{Image-Based Fact Verification}

The main function (Algorithm \ref{algorithm:factverification}) orchestrates the entire process by calling the forgery identification algorithm (Algorithm \ref{algorithm:forgeryidentification}) and fact retrieval algorithm (Algorithm \ref{algorithm:factretrieval}). It takes the input image and outputs either a confirmation that the image is \textbf{authentic} or, if it is identified as forged, \textbf{the set of original images} from which the input image was derived.

\begin{algorithm}[H]
\captionsetup{font=scriptsize}
\caption{ImageBasedFactVerification}
\label{algorithm:factverification}
\scriptsize
\begin{algorithmic}[1]
\Function{ImageBasedFactVerification}{InputImage}
    \State (ForgeryFlag, ForgeryType, ForgeryMask) $\gets$ \textsc{ForgeryIdentification}(InputImage)
    \If{ForgeryFlag = \text{False}}
        \State \Return "The image is authentic"
    \EndIf
    \State OriginalImages $\gets$ \textsc{FactRetrieval}(InputImage, ForgeryType, ForgeryMask)
    \State \Return OriginalImages
\EndFunction
\end{algorithmic}
\end{algorithm}

\subsubsection{Forgery Identification}

The forgery identification algorithm determines if the image is forged, identifies the type of forgery, and predicts the forgery locations if necessary. This involves three primary steps: binary classification (Algorithm \ref{algorithm:binaryclassification}), forgery type classification (Algorithm \ref{algorithm:multiclassification}), and forgery segmentation (Algorithm \ref{algorithm:forgerydetection}).

\begin{algorithm}[H]
\captionsetup{font=scriptsize}
\caption{ForgeryIdentification}
\label{algorithm:forgeryidentification}
\scriptsize
\begin{algorithmic}[1]
\Function{ForgeryIdentification}{Image}
    \State ForgeryFlag $\gets$ \textsc{BinaryClassification}(Image)
    \If{ForgeryFlag = \text{False}}
        \State \Return (ForgeryFlag, \text{None}, \text{None})
    \EndIf
    \State ForgeryType $\gets$ \textsc{ForgeryTypeClassification}(Image)
    \If{ForgeryType $\in$ \{\text{`copy-move'}, \text{`image splicing'}\}}
        \State ForgeryMask $\gets$ \textsc{ForgerySegmentation}(Image, ForgeryType)
    \Else
        \State ForgeryMask $\gets$ \text{None}
    \EndIf
    \State \Return (ForgeryFlag, ForgeryType, ForgeryMask)
\EndFunction
\end{algorithmic}
\end{algorithm}

\textbf{Binary Classification}: The binary classification step involves determining whether the image is authentic or forged. This is achieved through a pre-trained binary classification model (Algorithm \ref{algorithm:authenticity}), which outputs a Boolean value indicating the authenticity of the image.

\begin{algorithm}[H]
\captionsetup{font=scriptsize}
\caption{BinaryClassification}
\label{algorithm:binaryclassification}
\scriptsize
\begin{algorithmic}[1]
\Function{BinaryClassification}{Image}
    \State ForgeryFlag $\gets$ \textsc{PredictForgery}(Image)
    \State \Return ForgeryFlag
\EndFunction
\end{algorithmic}
\end{algorithm}

\textbf{Forgery Type Classification}: Once an image is identified as forged, the next step is to classify the type of forgery. The forgery type classification function uses a pre-trained model (Algorithm \ref{algorithm:forgerytype}) to categorize the forgery as either `copy-move’ or `image splicing.’

\begin{algorithm}[H]
\captionsetup{font=scriptsize}
\caption{ForgeryTypeClassification}
\label{algorithm:multiclassification}
\scriptsize
\begin{algorithmic}[1]
\Function{ForgeryTypeClassification}{Image}
    \State ForgeryType $\gets$ \textsc{PredictForgeryType}(Image)
    \State \Return ForgeryType
\EndFunction
\end{algorithmic}
\end{algorithm}

\textbf{Forgery Segmentation}: For images classified as `copy-move’ or `image splicing,’ the forgery segmentation function detects and segments the specific regions of the image with tampering (Algorithm \ref{algorithm:forgerymask}). This segmentation is crucial for the subsequent fact retrieval phase.

\begin{algorithm}[H]
\captionsetup{font=scriptsize}
\caption{ForgerySegmentation}
\label{algorithm:forgerydetection}
\scriptsize
\begin{algorithmic}[1]
\Function{ForgerySegmentation}{Image, ForgeryType}
    \State ForgeryMask $\gets$ \textsc{PredictForgeryMask}(Image, ForgeryType)
    \State \Return ForgeryMask
\EndFunction
\end{algorithmic}
\end{algorithm}

\subsubsection{Fact Retrieval}
The fact retrieval phase retrieves the original images from which the forged image was derived. Global retrieval (Algorithm \ref{algorithm:globalretrieval}) is always performed, and, if the forgery type is ‘copy-move’ or ‘image splicing,’ local retrieval (Algorithm \ref{algorithm:localretrieval}) is also performed.

\begin{algorithm}[H]
\captionsetup{font=scriptsize}
\caption{FactRetrieval}
\label{algorithm:factretrieval}
\scriptsize
\begin{algorithmic}[1]
\Function{FactRetrieval}{ForgedImage, ForgeryType, ForgeryMask}
    \State OriginalImages $\gets$ \textsc{GlobalRetrieval}(ForgedImage)
    \If{ForgeryType $\in$ \{\text{'copy-move'}, \text{'image splicing'}\}}
        \State Segments $\gets$ \textsc{ExtractSegments}(ForgedImage, ForgeryMask)
        \State SegmentOriginalImages $\gets$ \textsc{LocalRetrieval}(Segments)
        \State \text{Add} SegmentOriginalImages \text{to} OriginalImages
    \EndIf
    \State \Return OriginalImages
\EndFunction
\end{algorithmic}
\end{algorithm}

\textbf{Local Retrieval}: Local retrieval is used for forgery types that involve specific segments of the image, such as ‘copy-move’ or ‘image splicing.’ This function retrieves the original images (Algorithm \ref{algorithm:retrievaloriginalsegment}) of the segments cropped from the image in accordance with the detected mask (Algorithm \ref{algorithm:extractsegments}).

\begin{algorithm}[H]
\captionsetup{font=scriptsize}
\caption{LocalRetrieval}
\label{algorithm:localretrieval}
\scriptsize
\begin{algorithmic}[1]
\Function{LocalRetrieval}{Segments}
    \State OriginalImages $\gets$ [\ ]
    \For{each Segment in Segments}
        \State OriginalImage $\gets$ \textsc{RetrieveOriginal}(Segment)
        \State append OriginalImage to OriginalImages
    \EndFor
    \State \Return OriginalImages
\EndFunction
\end{algorithmic}
\end{algorithm}

\textbf{Global Retrieval}: Global retrieval is always used to retrieve the entire original image. This function retrieves the original image(s) (Algorithm \ref{algorithm:retrievaloriginalfull}) corresponding to the entire forged image.

\begin{algorithm}[H]
\captionsetup{font=scriptsize}
\caption{GlobalRetrieval}
\label{algorithm:globalretrieval}
\scriptsize
\begin{algorithmic}[1]
\Function{GlobalRetrieval}{Image}
    \State OriginalImages $\gets$ \textsc{RetrieveOriginal}(Image)
    \State \Return OriginalImages
\EndFunction
\end{algorithmic}
\end{algorithm}

\textbf{Extract Segments}: This function extracts the forgery segments from the image by using the provided mask. It processes the mask to identify and extract the relevant segments for local retrieval (Algorithm \ref{algorithm:masktosegments}).

\begin{algorithm}[H]
\captionsetup{font=scriptsize}
\caption{ExtractSegments}
\label{algorithm:extractsegments}
\scriptsize
\begin{algorithmic}[1]
\Function{ExtractSegments}{Image, Mask}
    \State Segments $\gets$ \textsc{ApplyMaskToExtractSegments}(Image, Mask)
    \State \Return Segments
\EndFunction
\end{algorithmic}
\end{algorithm}

\subsection{Utility Functions}
Several utility functions are used within the main and sub-algorithms to predict forgeries, classify forgery types, segment forgery regions, and retrieve original images. These functions abstract the model predictions and image processing steps.

\subsubsection{Predict Forgery}
The `PredictForgery' function implements the binary classification model to determine if the image is forged.

\begin{algorithm}[H]
\captionsetup{font=scriptsize}
\caption{PredictForgery}
\label{algorithm:authenticity}
\scriptsize
\begin{algorithmic}[1]
\Function{PredictForgery}{Image}
    \State \Comment{Implementation of the binary classification model}
    \State \Return \text{Boolean value indicating whether the image is forged}
\EndFunction
\end{algorithmic}
\end{algorithm}

\subsubsection{Predict Forgery Type}
The `PredictForgeryType' function implements the forgery type classification model used to identify the type of forgery.

\begin{algorithm}[H]
\captionsetup{font=scriptsize}
\caption{PredictForgeryType}
\label{algorithm:forgerytype}
\scriptsize
\begin{algorithmic}[1]
\Function{PredictForgeryType}{Image}
    \State \Comment{Implementation of the forgery type classification model}
    \State \Return \text{Type of forgery (e.g., `copy-move', `image splicing')}
\EndFunction
\end{algorithmic}
\end{algorithm}

\subsubsection{Predict Forgery Mask}
The `PredictForgeryMask' function implements the forgery segmentation model used to identify the tampered regions.

\begin{algorithm}[H]
\captionsetup{font=scriptsize}
\caption{PredictForgeryMask}
\label{algorithm:forgerymask}
\scriptsize
\begin{algorithmic}[1]
\Function{PredictForgeryMask}{Image, ForgeryType}
    \State \Comment{Implementation of the forgery segmentation model}
    \State \Return \text{Mask indicating the forgery region}
\EndFunction
\end{algorithmic}
\end{algorithm}

\subsubsection{Retrieve Original}
The `RetrieveOriginal' function retrieves the original image corresponding to a given segment or the entire forged image.

\begin{algorithm}[H]
\captionsetup{font=scriptsize}
\caption{RetrieveOriginal (Segment)}
\label{algorithm:retrievaloriginalsegment}
\scriptsize
\begin{algorithmic}[1]
\Function{RetrieveOriginal}{Segment}
    \State \Comment{Implementation of the retrieval model}
    \State \Return \text{Original image corresponding to the segment}
\EndFunction
\end{algorithmic}
\captionsetup{font=scriptsize}
\caption{RetrieveOriginal (Full-Image)}
\label{algorithm:retrievaloriginalfull}
\scriptsize
\begin{algorithmic}[1]
\Function{RetrieveOriginal}{Image}
    \State \Comment{Implementation of the retrieval model}
    \State \Return \text{Original image corresponding to the entire image}
\EndFunction
\end{algorithmic}
\end{algorithm}

\subsubsection{Apply Mask to Extract Segments}
The `ApplyMaskToExtractSegments' function processes the mask to identify and extract relevant segments from the image.

\begin{algorithm}[H]
\captionsetup{font=scriptsize}
\caption{ApplyMaskToExtractSegments}
\label{algorithm:masktosegments}
\scriptsize
\begin{algorithmic}[1]
\Function{ApplyMaskToExtractSegments}{Image, Mask}
    \State \Comment{Implementation of the mask application process}
    \State \Return \text{Extracted segments from the image}
\EndFunction
\end{algorithmic}
\end{algorithm}

\section{Effect of Proportion of Forgery Parts}
\label{sec:proportionimpacts}

The size of the manipulated regions within images significantly affects the effectiveness of forgery detection algorithms and the interpretability of detection outcomes. Unlike conventional forgery datasets, our dataset deliberately increases the average proportion of forgery parts to simulate real-world scenarios and enhance the difficulty of detection.

\begin{figure}[h]
    \begin{center} 
        \includegraphics[width=1\linewidth]{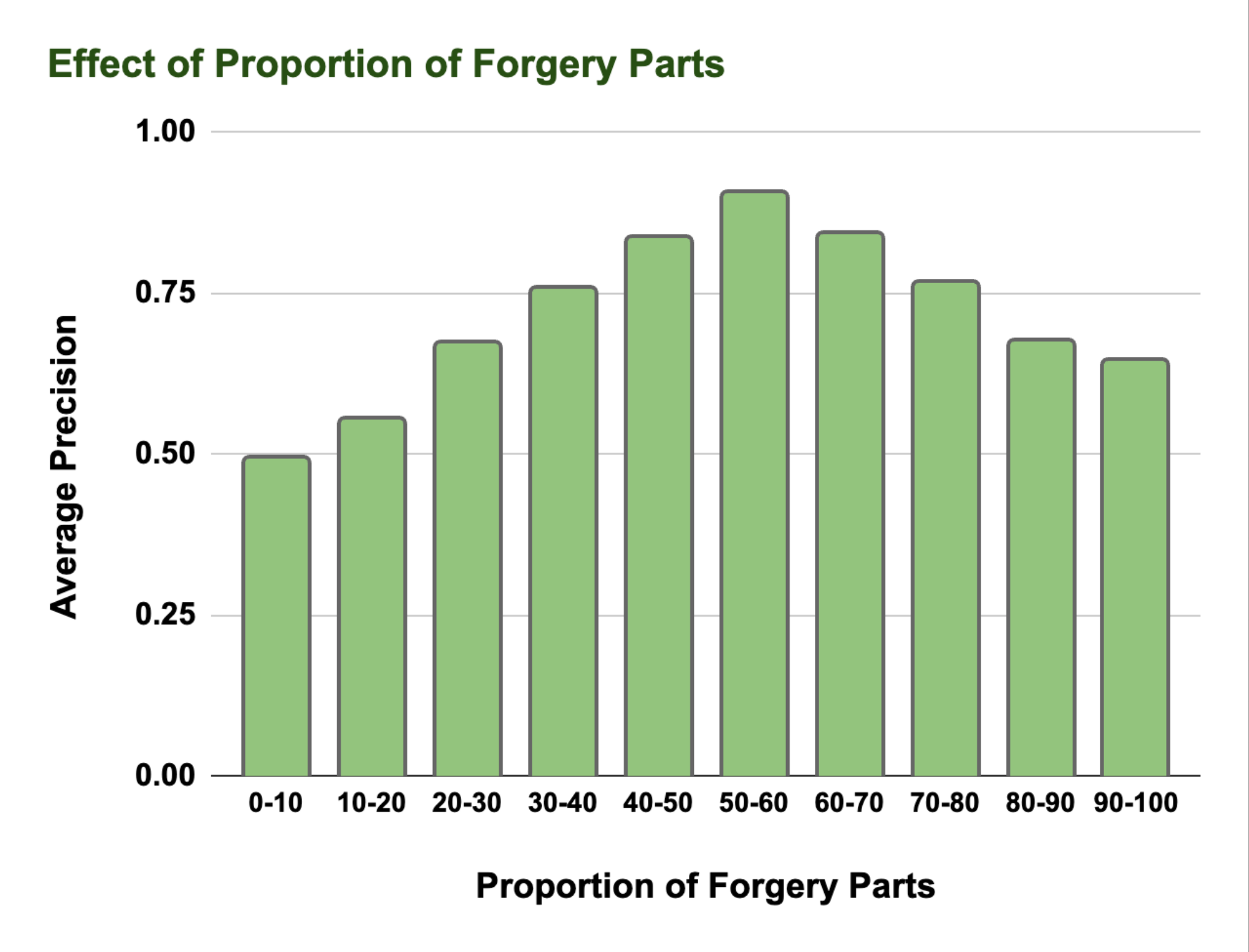}
    \end{center}
    \caption{Average precision performance of fact verification as the proportion of forgery parts was gradually increased from 0\% to 100\%.}
    \label{fig:forgery-size}
\end{figure}

To assess the effect of the proportion of forgery parts on verification performance, we divided queries into ten groups on the basis of the proportion and evaluated each group using our two-phase framework. Figure \ref{fig:forgery-size} illustrates the average precision for each group.

Analysis of the results revealed that accuracy peaked when the proportion ranged between $50$\% and $60$\%. This observation aligns with our intuition: an excessively small or large proportion poses challenges in identifying original images corresponding to the forgery parts or background areas.

A critical parameter in forgery detection is the size of the forgery mask, which is critical in achieving a balance between inclusivity and precision. It will greatly affect the robustness and reliability of the forgery detection algorithms facing diverse manipulative techniques and image characteristics.

\appendices

\bibliographystyle{IEEEtran}